\documentclass[conference]{IEEEtran}
\IEEEoverridecommandlockouts
\usepackage{cite}
\usepackage{amsmath,amssymb,amsfonts}
\usepackage{algorithmic}
\usepackage{graphicx,booktabs,multirow}
\usepackage{textcomp}
\usepackage{xcolor, soul}
\usepackage{enumitem}

\usepackage{url}
\usepackage{fancyvrb,fancyhdr}
\usepackage{commath}
\usepackage{comment}
\usepackage{fleqn,multirow,wrapfig,lineno,hyperref}
\hypersetup{
    colorlinks=true,
    linkcolor=blue,
    filecolor=blue,      
    urlcolor=blue,
    bookmarksopen=true,
    citebordercolor=blue,
    filebordercolor=blue,
    linkbordercolor=blue,
    menubordercolor=blue,
    urlbordercolor=blue
}
\sethlcolor{yellow}

\definecolor{bronze}{rgb}{0.8, 0.5, 0.2}
\definecolor{blue}{rgb}{0, 0, 1}
\definecolor{green}{rgb}{0, 1, 0}
\definecolor{black}{rgb}{0, 0, 0}
\definecolor{red}{rgb}{1, 0, 0}

\newcommand{\pn}[1]{{\color{black}#1}}

\newcommand{\myVspace}{\vspace{5 pt}}

\graphicspath{{figs/}}

\def\BibTeX{{\rm B\kern-.05em{\sc i\kern-.025em b}\kern-.08em
    T\kern-.1667em\lower.7ex\hbox{E}\kern-.125emX}}

\begin{document}


\title{
Efficient and Safe Contact-rich pHRI\\via Subtask Detection and Motion Estimation\\using Deep Learning
}

\author{Pouya~P.~Niaz$^{1}$,
        Engin~Erzin$^{2}$,
        and Cagatay~Basdogan$^{1}$
\thanks{$^{1}$ P. P. Niaz and C. Basdogan (\textit{corresponding author}) are with the Robotics and Mechatronics Laboratory (RML) and the KUIS AI Center, Koc University, Sariyer, Istanbul 34450, Turkey.
{\tt\small {\{pniaz20, cbasdogan\}@ku.edu.tr}}}
\thanks{$^{2}$ E. Erzin is with the Multimedia, Vision, and Graphics Laboratory and the KUIS AI Center, Koc University, Sariyer, Istanbul 34450, Turkey. {\tt\small{eerzin@ku.edu.tr}}}}

\maketitle

\begin{abstract}
This paper proposes an adaptive admittance controller for improving efficiency and safety in physical human-robot interaction (pHRI) tasks in small-batch manufacturing that involve contact with stiff environments, such as drilling, polishing, cutting, etc. We aim to minimize human effort and task completion time while maximizing precision and stability during the contact of the machine tool attached to the robot's end-effector with the workpiece. To this end, a two-layered learning-based human intention recognition mechanism is proposed, utilizing only the kinematic and kinetic data from the robot and two force sensors. A ``subtask detector" recognizes the human intent by estimating which phase of the task is being performed, e.g., \textit{Idle}, \textit{Tool-Attachment}, \textit{Driving}, and \textit{Contact}. Simultaneously, a ``motion estimator" continuously quantifies intent more precisely during the \textit{Driving} to predict when \textit{Contact} will begin. The controller is adapted online according to the subtask while allowing early adaptation before the \textit{Contact} to maximize precision and safety and prevent potential instabilities. Three sets of pHRI experiments were performed with multiple subjects under various conditions. Spring compression experiments were performed in virtual environments to train the data-driven models and validate the proposed adaptive system, and drilling experiments were performed in the physical world to test the proposed methods' efficacy in real-life scenarios. Experimental results show subtask classification accuracy of 84\% and motion estimation R\textsuperscript{2} score of 0.96. Furthermore, 57\% lower human effort was achieved during \textit{Driving} as well as 53\% lower oscillation amplitude at \textit{Contact} as a result of the proposed system.
\end{abstract}


\section{Introduction}
    \label{sec:introduction}

Many pHRI tasks in small-batch manufacturing involving contacts of the machine tool with the environment, such as collaborative drilling, polishing, cutting, etc., contain different phases (sub-tasks) with different requirements, rendering an interaction controller, such as an admittance controller, with fixed parameters impractical. For instance, in the \textit{Driving} phase, where a human is guiding the machine tool attached to the robot in free space by hand, low admittance damping is desired to minimize human effort since the human intention is to bring the machine tool close to the workpiece with minimal resistance from the robot. By contrast, the \textit{Contact} phase, which involves contact of the machine tool with a workpiece, requires higher admittance damping to ensure stability~\cite{Aydin2018, Aydin2021} since the human intends to complete the task safely. It is crucial to emphasize that detecting the \textit{Contact} phase before its occurrence is essential for smoothly increasing the admittance damping. This necessitates the estimation of human motion intent during the \textit{Driving} phase. Delayed activation of the damping mechanism can lead to instability during the \textit{Contact} phase, whereas premature adaptation may inadvertently heighten human effort and extend the duration of the \textit{Driving} phase, ultimately prolonging the task's total completion time. 

The preceding discussion underscores the significance of accurately discerning human intent in physical and human-robot interaction (pHRI). It ensures the seamless transition between task phases, helps to optimize human-robot interaction by dynamically adjusting control parameters, and maintains system stability and efficiency—ultimately fostering a safer, more intuitive, and more efficient collaboration. 

Researchers have used various methods to detect human intent to improve task performance in pHRI systems. In controlled settings with known and predefined constraints, threshold values of velocity and force (or their derivatives) have been used for detecting human intent for adaptive pHRI ~\cite{Kucukyilmaz2013, Aydin2014, Sirintuna2020, Hamad2021, Madani2022}. Though easy to implement, such rule-based methods often fail in dynamic settings with changing conditions, which is why data-driven methods have started to replace them, owing to their ability to extract meaningful relationships and latent features. In tasks that involve reaching and co-manipulation, minimum-jerk models are often employed to infer human motion intentions, thereby enabling the robot to assist users in completing tasks more efficiently and with reduced effort \cite{Corteville2007, Zhao2021}. However, complex pHRI tasks rarely follow minimum-jerk profiles, limiting their practicality. Alternatively, Hidden Markov Models (HMM) have been used in pHRI tasks with applications in collaborative assembly, to produce collision-free movements, detect human intentions and assist them accordingly~\cite{Wang2020b, Lin2022}. HMM processes the latest state of the system and may be ineffective when long-term dependencies affect future motions and behaviors significantly. In addition to the approaches above, shape-invariant representations~\cite{Vochten2016, Vochten2019}, probabilistic movement primitives~\cite{Paraschos2018}, and parameterized probabilistic principal component analysis~\cite{Perico2020} have been proposed for modeling human behavior in pHRI tasks ~\cite{Ravichandar2020, Burlizzi2022}. While simple to implement and able to model stochasticity, these methods often do not capture all the complex relationships between kinematic/kinetic time series signals and human behavior in contact-rich pHRI tasks. Other data-driven techniques such as Dynamic Time Warping (DTW), Gaussian Process Regression (GPR), and Random Forests (RF) have also been used for intention recognition or motion prediction in pHRI~\cite{Maeda2017, Alsaadi2021, Wang2022}. Once again, such methods can work effectively in simple tasks but may fail to extract deeper latent features in complex ones. Deep Learning (DL) has shown promising results in time series prediction and sequence modeling due to its ability to capture deeper representations and learn nonlinear relationships. It has been successfully used for intention recognition, motion prediction, and human behavior modeling in pHRI~\cite{Mazhar2018, Maithani2019, Sirintuna2020a, Liu2021, Atkins2023}.

\begin{figure}[t]
    \myVspace
	\includegraphics[width=\columnwidth]{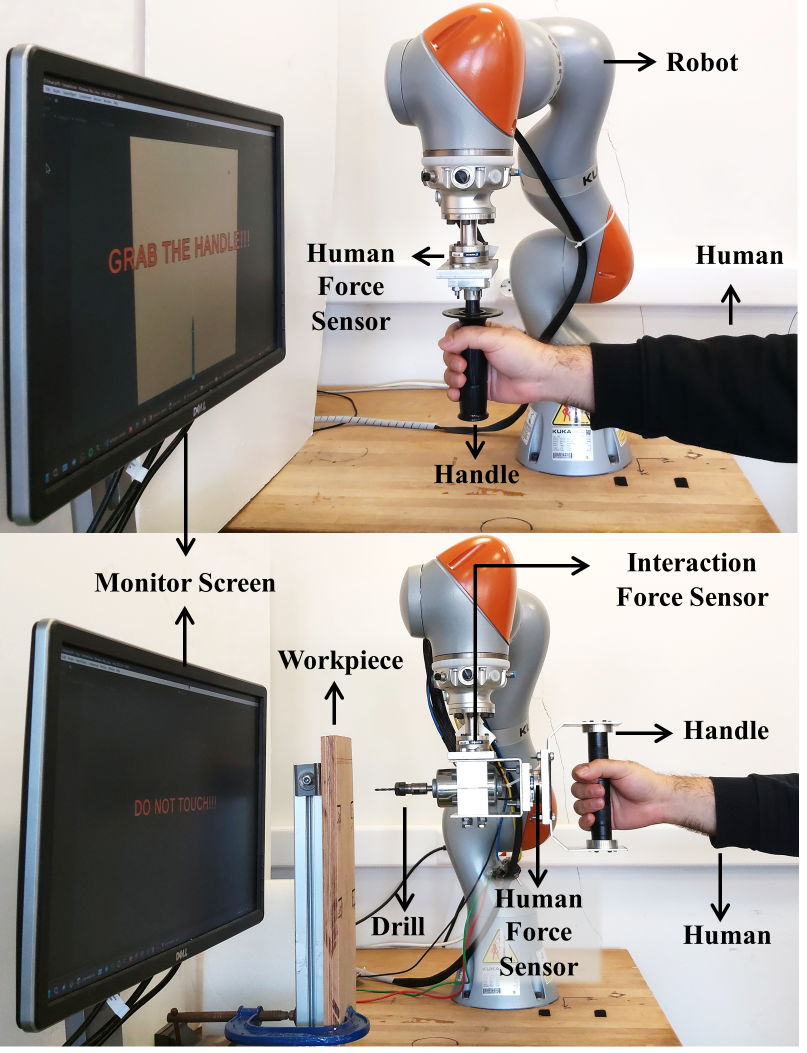}
	\centering
	\caption{Hardware setup in the virtual spring compression task (top) and the real drilling task (bottom).}
	\label{fig:setup}
\end{figure}

In this study, a human and a robot (with a machine tool attached to its end-effector and equipped with an admittance controller) were envisioned executing a collaborative drilling task (Fig.~\ref{fig:setup}), which involves contact with a stiff workpiece. The task was divided into separate subtasks, each with its own set of desired/required control parameters, thereby needing an adaptive system for better task efficiency. To this end, a \textbf{subtask detector} recognized the current subtask and adapted the admittance controller at subtask transitions as proposed in our earlier study~\cite{Guler2022}. Realistically, however, a subtask detector only ever reacts to changes in system behavior at subtask transitions with inevitable delay, \textit{after} they occur.
Hence, in a task like drilling, \textit{Contact} begins with low damping values remaining from the \textit{Driving}. The stiff contact instantly leads to instability as in oscillation or bounce-back \textit{before} the subtask detector recognizes \textit{Contact} and adapts the controller. This renders the subtask detector less effective in preventing instability at \textit{Contact}. Higher damping in \textit{Driving} would mitigate the problem but instead fail to minimize human effort, nullifying the advantage of adaptive control. The goal is to choose low damping for \textit{Driving} and high damping for \textit{Contact} while maintaining stability regardless of material stiffness and contact rigidness. An additional \textbf{motion estimator} is thus proposed in this study, so that the occurrence of \textit{Contact} can be anticipated to adapt the controller earlier in time. As a result, as reported in the experimental results, human effort was minimized during \textit{Driving}, and stability was maintained throughout \textit{Contact}. We conduct collaborative drilling experiments both in virtual and real worlds to validate the efficacy of the proposed system. 

The contributions of this work are as follows:

\begin{itemize}[leftmargin=*]
     \item A two-layered ML architecture consisting of a \textbf{subtask detector} and a \textbf{motion estimator} was proposed for adapting the damping coefficient of an admittance controller in contact-rich pHRI tasks to minimize effort and maximize stability. Different ML models were explored and compared to determine which one works best. For this purpose, we utilized standard metrics and also introduced new ones to quantify the performance of \textbf{subtask detector} and \textbf{motion estimator}. The experimental results show that the proposed two-layer approach utilizing DL models for both layers led to better performance than the others. Moreover, the proposed two-layer ML approach reduces the instabilities at contact by 53\% compared to the \textbf{subtask detector} alone, as investigated by Guler et al. \cite{Guler2022}. This outcome shows the add-on value of the \textbf{motion estimator} in adapting the admittance damping during the transitioning from \textit{Driving} to \textit{Contact}.  
    \item We conducted spring compression experiments in virtual environments (VEs), imitating drilling in physical world, to train different ML models and compare their performances. These experiments were performed with multiple human subjects under various conditions, with and without the adaptive admittance controller. Our study highlights the benefits of VEs for training an ML model, an approach that aligns well with the concept of 'Sim2Real'—simulation to reality. Since we successfully updated the control loop of our pHRI system at a high rate of 500 Hz, we could render contact forces and display real-time haptic feedback to a user during the simulations in VEs. This approach eliminated the necessity for extensive physical world trials, which is time-consuming and not practical for the drilling task. After the two-layer architecture was rigorously trained and validated through simulations in VEs, it was implemented in the physical world, where its effectiveness was tested in real-life scenarios through a series of actual drilling experiments. The results show that the subtask classifier can achieve an accuracy of 80\%, while the motion estimator could estimate the progress with an R\textsuperscript{2} value of 0.95 in both virtual spring compression and real drilling experiments.
        
\end{itemize}

In Section~\ref{sec:approach}, our hardware setup, adaptive control system, and evaluation metrics are explained. Section~\ref{sec:expA} explains the spring compression experiments performed in VEs for training the ML models offline, while Section~\ref{sec:expB} reports the results of online deployment and testing of the models in the same environment. Section~\ref{sec:expC} reports real drilling experiments' subtask detection and motion estimation performance while deploying the ML models online. 
Section~\ref{sec:conclusion} concludes the study.

\section{Approach}
    \label{sec:approach}

\subsection{Hardware Setup}
    \label{sec:hardware}

In this study, two types of experiments were performed.
Spring compression experiments were performed in a virtual environment (VE) for offline training and online testing of ML models.
Drilling experiments were performed in the physical world for final deployment and assessment. Fig.~\ref{fig:setup} shows the hardware components of our experimental setup for both environments.

The virtual spring compression setup included a cobot (LBR iiwa 7, KUKA Inc.), a handle for the human to grab, and a monitor for visual feedback. A force sensor (Mini 45, ATI Inc.) connected the handle to the robot's end-effector to measure human forces. The real drilling setup included a DC motor, a drill bit, and another force sensor (Mini 45, ATI Inc.) to measure the interaction forces applied to the end-effector.

\subsection{Closed Loop Control Architecture}
    \label{sec:controlsystem}

\begin{figure}[t]
    \myVspace
	\includegraphics[width=\columnwidth]{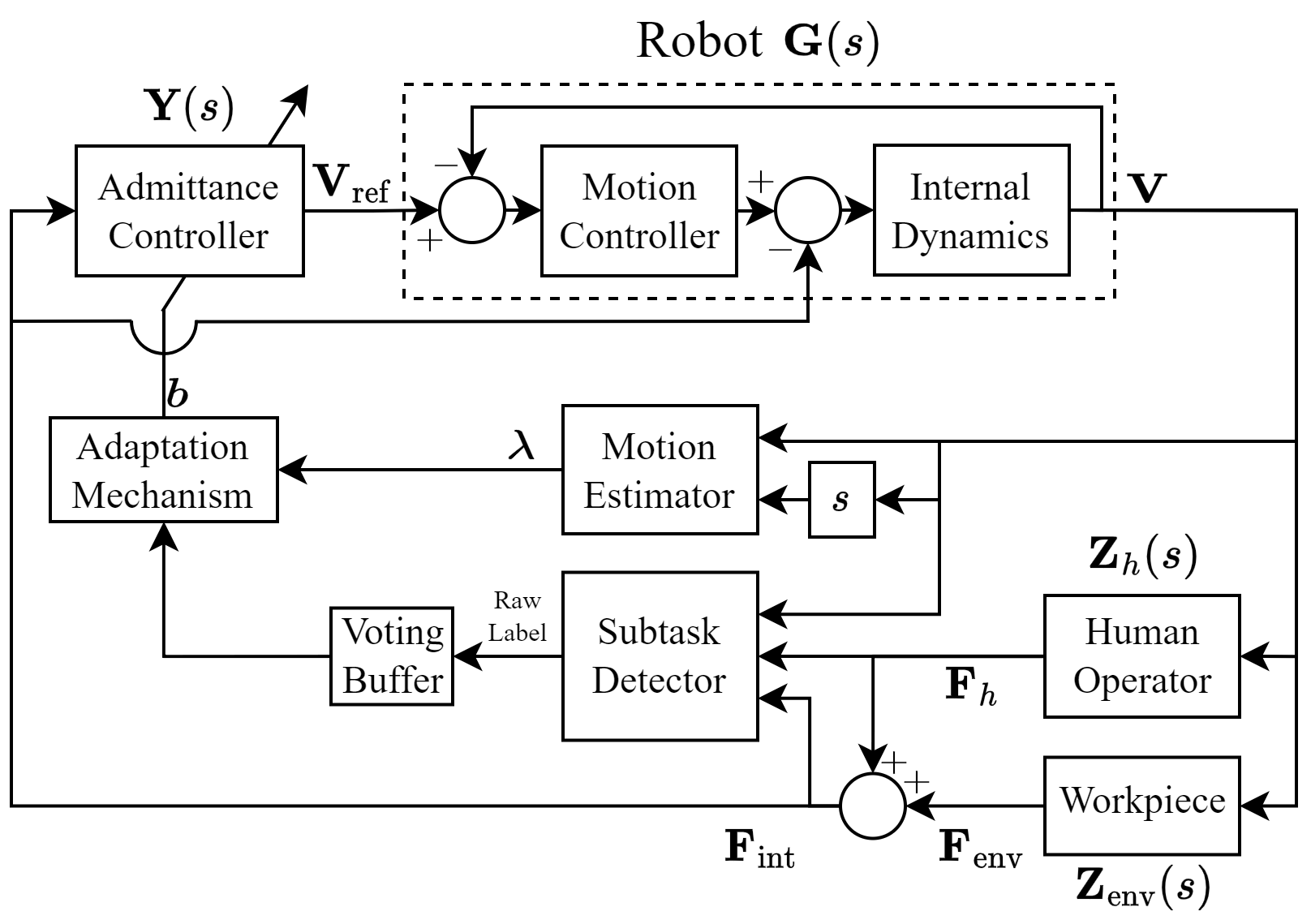}
	\centering
	\caption{The closed-loop control architecture used in our study. All variables are 3-dimensional vectors, corresponding to 3 translational degrees of freedom in the Cartesian space.}
	\label{fig:controlsystem}
\end{figure}

Our control loop for pHRI, updated at 500 Hz, is shown in Fig.~\ref{fig:controlsystem}.
Accordingly, $\mathbf{F}_h$ and $\mathbf{F}_\mathrm{env}$ are human and environment forces, respectively, and $\mathbf{F}_\mathrm{int} = \mathbf{F}_h + \mathbf{F}_\mathrm{env}$ is the interaction force. $\mathbf{V}_\mathrm{ref}$ is the reference velocity, and $\mathbf{V}$ is the real velocity.
%
Additionally, $\mathbf{Z}_h(s)$ and $\mathbf{Z}_\mathrm{env}(s)$ are the unknown human and environment (workpiece) mechanical impedances, respectively. The admittance controller allows three translational degrees of freedom, for each of which the admittance controller has the following transfer function:

\begin{equation}
    Y(s) = \frac{V_\mathrm{ref}(s)}{F_\mathrm{int}(s)} = \frac{1}{m\,s+b}
    \label{eq:admittancectrl}
\end{equation}
where $m$~[kg] and $b$~[Ns/m] are the admittance mass and damping parameters, respectively, chosen to be equal among all 3 DOFs \pn{to maintain consistency when movement directions, target orientations, and material properties changed.}

\subsection{Adaptation Mechanism}
    \label{sec:adaptivecontrol}

\begin{paragraph}{\textbf{Subtask Detector}} 
We generically divided contact-rich pHRI tasks, such as drilling, polishing, cutting, etc., into four subtasks:
\begin{enumerate}
    \item \textit{Idle}: The robot is stationary, and there is no input.
    \item \textit{Tool-Attachment}: The human holds the handle to attach the drill bit, polishing pad, saw band, etc.
    \item \textit{Driving}: The human guides the robot in free space towards the workpiece.
    \item \textit{Contact}: The human and the robot perform drilling, polishing, cutting, etc., on the workpiece.
\end{enumerate}
In our two-layer approach, the subtask detector shown in Fig.~\ref{fig:controlsystem} was always active. It took sliding windows of magnitudes $||\mathbf{V}||$, $||\mathbf{F}_\mathrm{int}||$ and $||\mathbf{F}_h||$ as the inputs to predict the current subtask. The ``voting buffer" in Fig.~\ref{fig:controlsystem} took the most frequent prediction in the last 100 steps of the control loop (i.e., 0.2 seconds).
\end{paragraph}

\begin{figure}[t]
	\includegraphics[width=0.8\columnwidth]{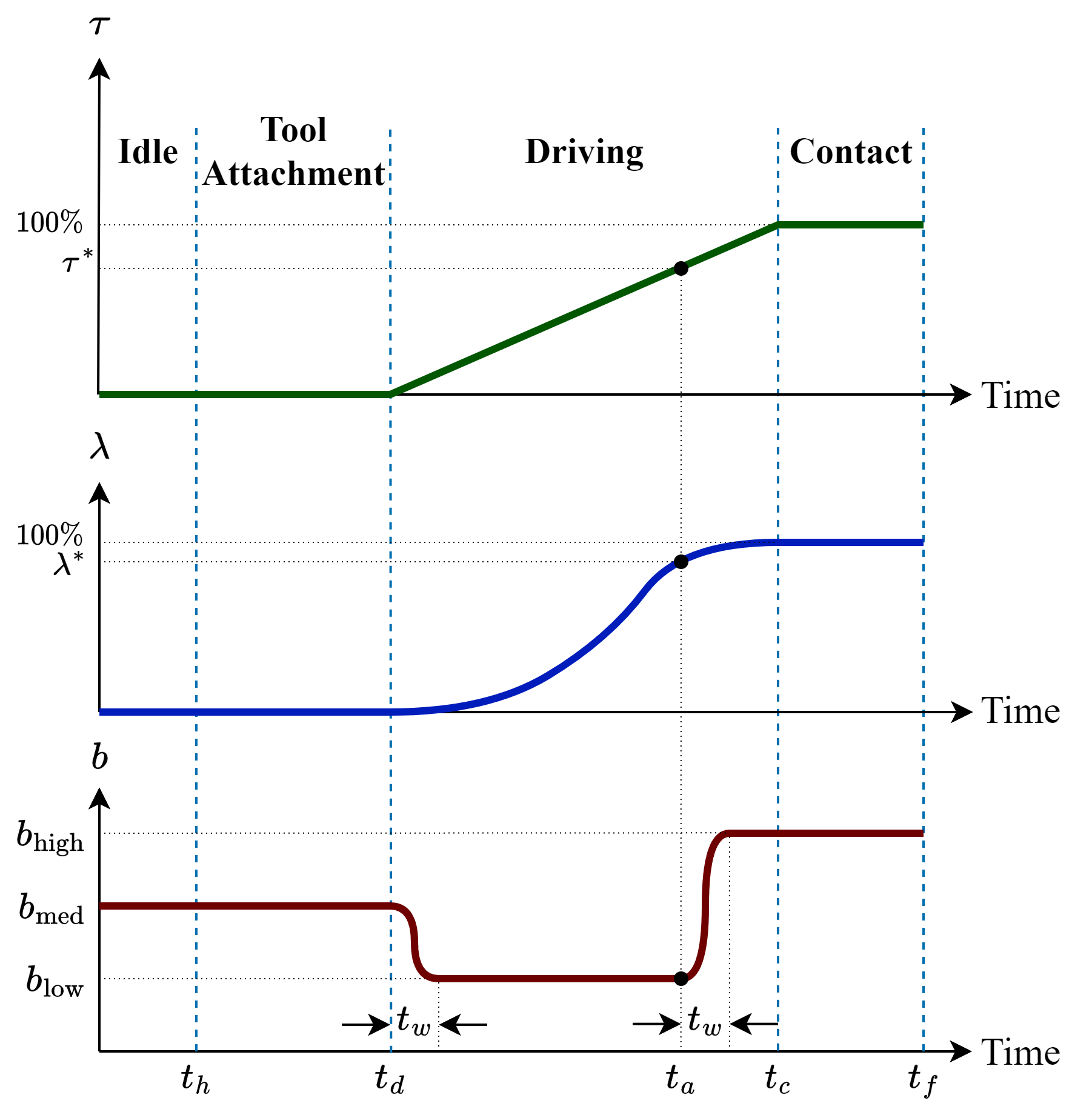}
	\centering
	\caption{Time progress $\tau$, trajectory progress $\lambda$ and admittance damping $b$ vs. time $t$ across all subtasks. Damping is adapted at $t_d$ when \textit{Driving} begins, and at $t_a$ when the estimated progress reaches the adaptation threshold, i.e., $\tau=\tau^*$, or $\lambda=\lambda^*$, just before \textit{Contact} occurs at $t_c$.
    The trial ends when the drilling/spring compression reaches its desired depth of 4~[mm] at $t_f$.
    }
	\label{fig:adaptplot}
\end{figure}

\begin{paragraph}{\textbf{Motion Estimator}}
The \textbf{motion estimator} shown in Fig.~\ref{fig:controlsystem} is only active during the \textit{Driving} phase. It takes sliding window sequences of velocity and acceleration signals, and estimates the progress, which starts with 0.0\% when \textit{Driving} begins and ends with 100\% at \textit{Contact}. We propose two metrics to track the human's behavior during the \textit{Driving} phase. We track (1) the ``time progress" $\tau$, which is the portion of the \textit{Driving} duration that has been traveled so far, and linearly increases in time, and (2) the ``trajectory progress" $\lambda$, which is the portion of the \textit{Driving} trajectory in length that has been traveled so far. It monotonically increases in time during \textit{Driving}, regardless of the shape of the spatial trajectory (see Fig.~\ref{fig:adaptplot}).

To clarify, the metrics for motion estimation are calculated as follows during the \textit{Driving} phase:
\begin{equation}
    \tau = \frac{t_d^\prime}{T_d} = \frac{t - t_d}{t_c - t_d}
    \label{eq:timeprogress}
\end{equation}
\begin{equation}
    \lambda = \frac{l_d}{L_d} = \frac{
    \int_{t_d}^{t}   \| \mathbf{V} \| \,dt
    }{
    \int_{t_d}^{t_c} \| \mathbf{V} \| \,dt
    }
    \label{eq:trajprogress}
\end{equation}
where $t_d^\prime$ is time since the beginning of \textit{Driving} ($t_d$), and $T_d$ is the time duration of \textit{Driving}. Also, $l_d$ is the length of the currently traveled trajectory in \textit{Driving}, and $L_d$ is the total trajectory length in \textit{Driving}.
\end{paragraph}

The adaptation mechanism is summarized in Fig.~\ref{fig:adaptplot}. Accordingly, the admittance damping $b$ in~\eqref{eq:admittancectrl} was adapted for every degree of freedom while the admittance mass was kept constant, $m=50\,\text{[kg]}$. 
A low damping $b_\mathrm{low}=100\,\text{[Ns/m]}$ was chosen for the \textit{Driving} to minimize human effort, while a higher damping $b_\mathrm{high}=500\,\text{[Ns/m]}$ was chosen for \textit{Contact} to maximize stability.
To prevent jerky motion at the beginning of \textit{Driving} and make the robot slightly more stable during \textit{Idle} and \textit{Tool-Attachment}, a medium value of admittance damping $b_\mathrm{med}=300\,\text{[Ns/m]}$ was chosen. 
As shown in Fig.~\ref{fig:adaptplot}, a progress value close to 100\% called ``Adaptation Threshold" was chosen (denoted as $\tau^*$ and $\lambda^*$ for the time elapsed and the distance traveled within the \textit{Driving}, respectively), so when the estimated progress reached it at time $t=t_a$, it was assumed that \textit{Driving} is close to the end and \textit{Contact} was imminent, so damping was adjusted for \textit{Contact}. Damping adaptation was implemented by using a cubic polynomial over a brief duration of $t_w=200,\text{ms}$ to achieve a smooth profile.

\subsection{Subtask Detection Methods}

Subtask detection is a time series classification problem, which can be treated in several ways. Since reliable classification performances are often required in real-time pHRI systems to adapt the system on time without delay, multilayer perceptrons were used in our previous study~\cite{Guler2022} for subtask detection. Since RNN and CNN architectures can benefit from weight-sharing and are more capable of extracting useful latent information from time series, LSTM models and CNN models with 1D convolutions were explored in this study.

\subsection{Motion Estimation Methods}

Several methods can be utilized for this regression problem. In this study, the minimum-jerk model, DTW, GPR, and deep learning ~\cite{Maeda2001, Lank2007, Zhao2021, Maeda2017, Li2020, Maithani2019, Sirintuna2020a} have been evaluated, considering each one's advantages and weaknesses. The models discussed below were trained on data collected in many experimental trials, as explained in Section~\ref{sec:experiments} later.



\subsubsection{{\textbf{Minimum-Jerk Model (MJ)}}}

A simple but popular model in motion estimation, the minimum-jerk trajectory is expressed as follows for each translational degree of freedom\cite{Hogan1984a, Flash1985}:
\begin{equation}
    p(t_d^\prime) = p_f\,\left(10\left(\frac{t_d^\prime}{T_d}\right)^3-15\left(\frac{t_d^\prime}{T_d}\right)^4+6\left(\frac{t_d^\prime}{T_d}\right)^5\right)
    \label{eq:minjerk}
\end{equation}
where $t_d^\prime$ is the elapsed time in \textit{Driving}, $p$ is position relative to the start point along either of the three axes and $p_f$
is its final value at \textit{Contact}.
In our implementation, the recorded trajectory in the \textit{Driving} phase was used at every instance with a least-squares curve fitting to estimate model parameters $p_f$ and $T_d$, which were used in trajectory and time progress estimation, respectively.
Though relatively simple and effective in point-to-point reaching motions, trajectories in complex pHRI scenarios may not follow the minimum-jerk trajectory.
This is the only method that required no training and was directly used during inference. The \texttt{least\textunderscore squares} module of the SciPy library~\cite{scipy} was used in Python for curve fitting.

\subsubsection{{\textbf{Dynamic Time Warping (DTW)}}}

A method for comparing complete time series more meaningfully than pure Euclidean distance~\cite{Callens2020, Ben2014}, DTW can also match a partial trajectory to a given complete template to estimate what portion of it best fits the currently recorded data. This ``portion" is the time progress. In our case, the template was calculated by interpolating the data at \textit{Driving} phase to a fixed length sequence for all trials and then averaging the sequences across the trials. Partial warping using DTW frequently exhibits suboptimal performance, particularly in scenarios where significant stochasticity and variability are present within the dataset, meaning not only the template may not be a good representation of all the data, but also partial observations may not match consistently well to different indices of the template~\cite{Callens2020, Ben2014}.
We used the \texttt{dtw-python} package for implementation~\cite{Tormene2009, Giorgino2009}.

\subsubsection{{\textbf{Gaussian Process Regression (GPR)}}}

GPR is a non-parametric statistical model with a closed-form solution. It uses kernel functions to model the covariance among features and predicts the target output's mean and variance based on the training data distribution.
Since GPR needs to store the kernel function outputs for the whole training set, it can be memory-intensive and impractical with large datasets.
In our implementation, concatenated feature vectors in a long sliding window during the \textit{Driving} phase were fed to the model at every time step of the control loop, which estimated the mean and variance of the progress value.
The Scikit-Learn library~\cite{Pedregosa2011} was used in Python for GPR implementation. Multiple kernel functions available in this package were tested and compared.
After much tuning, the Matern kernel~\cite{Rasmussen2006} was selected with an initial length scale of 1.0, length scale bounds of 10\textsuperscript{-20} to 10\textsuperscript{5}, and $\nu=0.5$. The global noise level was $\alpha=0.5$ for the GPR training function. The default optimizer was retained for optimizing the kernel parameters, and 20 restarts were allowed for it. The \texttt{GaussianProcessRegressor} class of the Scikit-Learn library~\cite{Pedregosa2011} in Python was used for implementation.

\subsubsection{{\textbf{Deep Learning (DL)}}}

\begin{figure}[t]
    \myVspace
	\includegraphics[width=0.8\columnwidth]{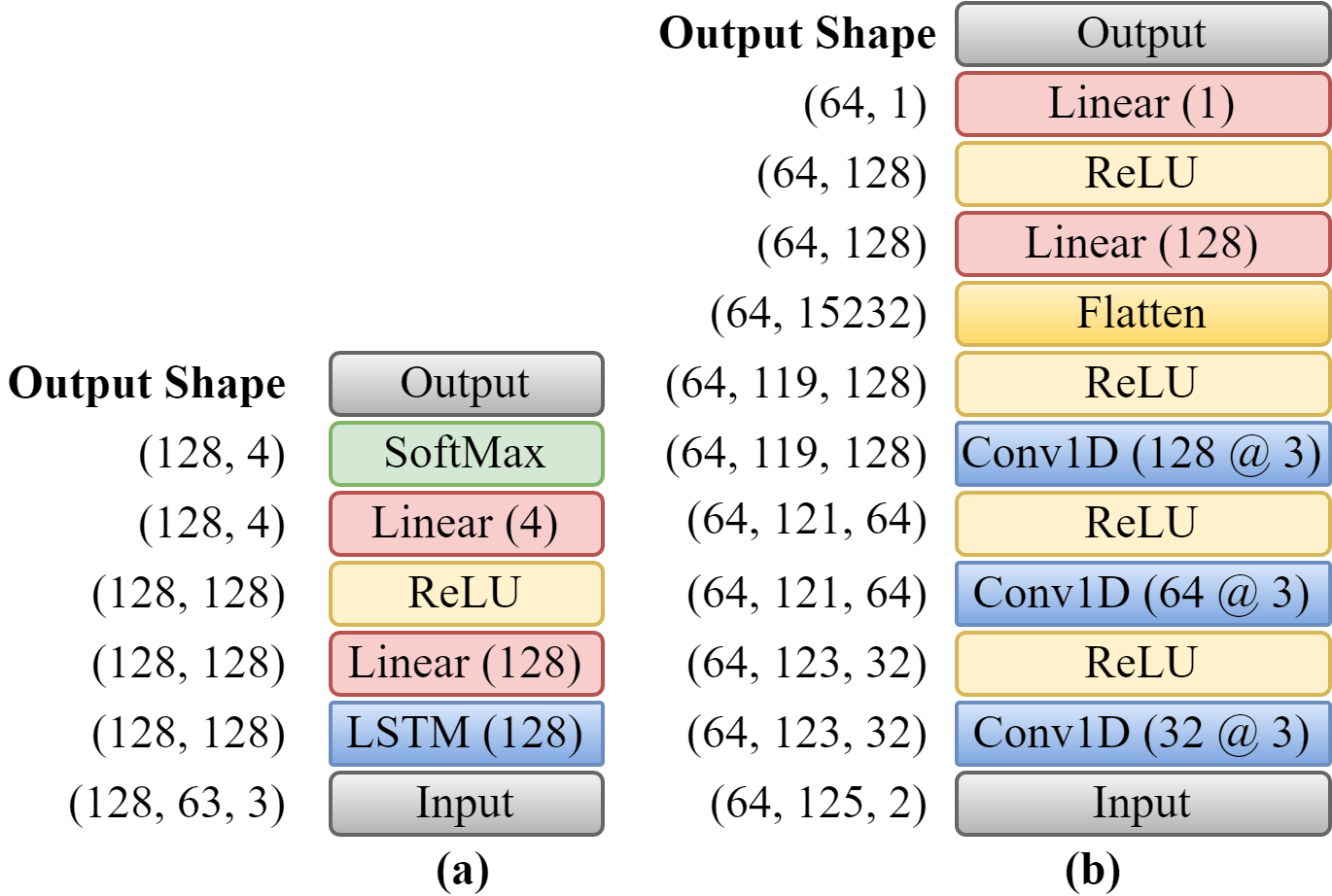}
	\centering
	\caption{Deep Learning architectures used in this study: (a) Subtask Detector LSTM, (b) Motion Estimator CNN. Numbers inside parentheses are hidden layer sizes for Linear layers, hidden sizes for LSTM layers, and the number of filters for Convolution layers, followed by filter sizes.}
    \label{fig:dlarch}
\end{figure}

In this study, Long Short-Term Memory (LSTM) networks, which are variants of Recurrent Neural Networks (RNN) were explored, along with Convolutional Neural Network (CNN) models with 1D convolution layers, which are specifically used for processing time series. A fixed-length sliding window acquired the data history within the \textit{Driving} phase and fed it to the neural network. The network output was always a single neuron without activation because it was a scalar regression problem. Vigorous regularization and hyperparameter optimization (HPO) were done to achieve the best generalization performance. The TensorFlow library~\cite{tensorflow} was used in Python for training and deploying the DL models. After extensive hyperparameter optimization, a Long Short-Term Memory (LSTM) model was chosen for subtask detection, and a Convolutional Neural Network (CNN) model with 1D convolutions was chosen for motion estimation. Fig.~\ref{fig:dlarch} shows the model architectures, while Table~\ref{tab:HPO} tabulates the hyperparameters chosen for these models. \pn{The data was sampled at 500 [Hz] and the sliding windows were pushed one step forward at each time step. The sequences extracted from the sliding windows were downsampled before being fed to the models to save time and memory during training and inference (See Table~\ref{tab:HPO}).}

\begin{table}[t]
\centering
\caption{DL model hyperparameters}
\label{tab:HPO}
\resizebox{\columnwidth}{!}{%
\begin{tabular}{lll}
\textbf{Parameter}                                                               & \textbf{Subtask Detection} & \textbf{Motion Estimation} \\ \hline
Sequence Length [sec]                                                            & 0.5                        & 8.0                        \\
Downsampling Rate                                                                & 4                          & 32                         \\
Num. Timesteps in Input                                                         & 63                         & 125                        \\
Num. Input Features                                                              & 3                          & 2                          \\
Num. Output Features                                                             & 4                          & 1                          \\
Minibatch Size                                                                   & 128                        & 64                         \\
Initial Learning Rate (LR)                                                           & 1.0 e-4                    & 1.0 e-4                    \\
LR Exponential Decay                                                             & 0.95                       & 0.95                       \\
Training Epochs                                                                  & 40                         & 40                         \\
Validation Set (\% Trainset)                                                     & 5                          & 5                          \\
Validation Patience (Epochs)                                                     & 10                         & 10                         \\
L1 Regularization                                                                & 0.0                        & 1.0 e-4                    \\
L2 Regularization                                                                & 0.0                        & 1.0 e-4                     
\end{tabular}%
}
\end{table}

\subsection{Evaluation Metrics}
    \label{sec:evaluationmetrics}

For all models and methods discussed above, a consistent evaluation method was used for their comparative analysis, both for subtask detection and motion estimation. The following metrics were calculated for every trial. Afterward, their statistics were extracted and compared.

\subsubsection{{\textbf{Subtask Detection Metrics}}}


\begin{itemize}
    \item {Classification metrics}: Model accuracy (ideally 100\%) and weighted F\textsubscript{1} scores (ideally 1.00).
    \item {Concurrency}: Model accuracy within transition intervals, i.e., a small time interval around subtask transitions (ideally 100\%). This quantifies the degree of lead or lag exhibited by the subtask detector in relation to actual subtask transitions. Lead and lag can result in premature and delayed adaptations, leading to undesirable or unstable interactions.
    \item {Fluctuation}: The amount of fluctuation in the predictions outside transition intervals. It is the mean rate at which the predicted subtask changes, expressed in the number of fluctuations per second, i.e., Hz (ideally 0.00 Hz). Fluctuations can cause unwanted changes and fluctuations in control parameters, leading to potentially unstable behavior.
\end{itemize}

\subsubsection{{\textbf{Motion Estimation Metrics}}}

These include:

\begin{itemize}
    \item {Regression metrics}: Root Mean Squared Error (RMSE, ideally 0.00) and R\textsuperscript{2} score (ideally 1.00).
    \item {Maximum Threshold}: The maximum allowable adaptation threshold leading to successful adaptation before \textit{Contact}, i.e., the maximum estimated progress throughout \textit{Driving} (ideally 100\%).
    \item {Mistiming}: Assuming the Maximum Threshold was chosen, the error in terms of normalized time (or trajectory length) between the Maximum Threshold and when adaptation occurs (ideally 0.00\%). This measures the lead or lag between the desired and true adaptation points.
\end{itemize}

\subsubsection{{\textbf{Task Performance Metrics}}}

After deploying the trained models on the adaptive control system and testing them in experiments, the performance of the proposed system needs to be evaluated in terms of the final objectives we are trying to achieve in this study. These include the following:

\begin{itemize}
    \item Average human force throughout \textit{Driving},
    \begin{equation}
        F_h^{\mathrm{ave}}=\frac{1}{T_d}\int_{t_d}^{t_c}\|\mathbf{F}_h(t)\|\,dt,\:\mathrm{[N]}
        \label{eq:Fhave}
    \end{equation}
    \item Average velocity throughout \textit{Driving},
    \begin{equation}
        V_{\mathrm{ave}} = \frac{1}{T_d}\int_{t_d}^{t_c}\|\mathbf{V}(t)\|\,dt,\:\mathrm{[m/s]}
        \label{eq:Vave}
    \end{equation}
    \item Total human effort throughout \textit{Driving},
    \begin{equation}
        E_h^{\mathrm{tot}} = \int_{t_d}^{t_c} \|\mathbf{F}_h(t)\|\,\|\mathbf{V}(t)\|\,dt,\:\mathrm{[J]}
        \label{eq:Ehtot}
    \end{equation}
    \item Oscillation of the end-effector velocity at \textit{Contact}, expressed as the maximum magnitude observed in its spectrogram, after detrending it with a high-order high-pass Butterworth filter with a cutoff frequency of 0.5 [Hz].
\end{itemize}

\subsection{Experiments}
    \label{sec:experiments}

A large amount of training data was required to train the DL models. Since extensive data collection with various distances and target locations/sizes would take a long time and consume significant material in a real drilling task, 
the experiments were conducted in virtual environments (VEs) using our collaborative robot and human subjects, where we simulated the workpiece as a linear spring with a coefficient of $K=8000\,\text{[N/m]}$.
We first conducted a virtual spring compression experiment with no motion estimation. The admittance damping was adapted based on hard-coded rules (Experiment A). Then, using the collected data, different models for subtask detection and motion estimation were trained and tuned \underline{offline}. Afterward, new experiments were performed to assess their \underline{online} performance. This time, the models were deployed in the adaptive control system, and adaptation was based on these trained models, not hard-coded rules (Experiment B). Eventually, to examine the real-life performance of these trained models, drilling experiments were also performed in physical world (Experiment C) with adaptation based on the trained models. Table~\ref{tab:experiments} summarizes the three experiments and their details.

\begin{table}[t]
\myVspace
\centering
\caption{Details of the experiments performed in this study}
\label{tab:experiments}
\resizebox{\columnwidth}{!}{%
\begin{tabular}{|c|ccc|}
\hline
\textbf{Designation} &
  \multicolumn{1}{c|}{A} &
  \multicolumn{1}{c|}{B} &
  C \\ \hline
\textbf{Motivation} &
  \multicolumn{1}{c|}{\begin{tabular}[c]{@{}c@{}}Collecting data\\ and training models\\ offline\end{tabular}} &
  \multicolumn{1}{c|}{\begin{tabular}[c]{@{}c@{}}Deploying trained\\ models online and\\ testing performance\end{tabular}} &
  \begin{tabular}[c]{@{}c@{}}Examining system\\ performance\\ in real tasks\end{tabular} \\ \hline
\textbf{Type} &
  \multicolumn{2}{c|}{Virtual Spring Compression} &
  Real Drilling \\ \hline
\textbf{Subjects} &
  \multicolumn{1}{c|}{5} &
  \multicolumn{1}{c|}{3} &
  3 \\ \hline
\textbf{\begin{tabular}[c]{@{}c@{}}Perpendicular\\ Distances\end{tabular}} &
  \multicolumn{1}{c|}{\begin{tabular}[c]{@{}c@{}}$L_p\in\{12,16,20,24\}[\text{cm}]$\end{tabular}} &
  \multicolumn{1}{c|}{$L_p=18\,\text{[cm]}$} &
  $L_p=18\,\text{[cm]}$ \\ \hline
\textbf{\begin{tabular}[c]{@{}c@{}}Target\\ Locations\end{tabular}} &
  \multicolumn{1}{c|}{4 corners} &
  \multicolumn{1}{c|}{4 corners} &
  4 corners \\ \hline
\textbf{\begin{tabular}[c]{@{}c@{}}Target\\ Sizes\end{tabular}} &
  \multicolumn{1}{c|}{\begin{tabular}[c]{@{}c@{}}$\text{IoD}\in\{3,5\}$\end{tabular}} &
  \multicolumn{1}{c|}{$\text{IoD}=4$} &
  $\text{IoD}=4$ \\ \hline
\textbf{\begin{tabular}[c]{@{}c@{}}Subtask\\ Detection\end{tabular}} &
  \multicolumn{1}{c|}{Hard-coded} &
  \multicolumn{2}{c|}{Trained Model} \\ \hline
\textbf{\begin{tabular}[c]{@{}c@{}}Motion\\ Estimation\end{tabular}} &
  \multicolumn{1}{c|}{N/A} &
  \multicolumn{2}{c|}{Trained Model} \\ \hline
\end{tabular}%
}
\end{table}

\section{Experiment A: Offline Training in VEs}
    \label{sec:expA}

\begin{figure}[t]
    \myVspace
	\includegraphics[width=\columnwidth]{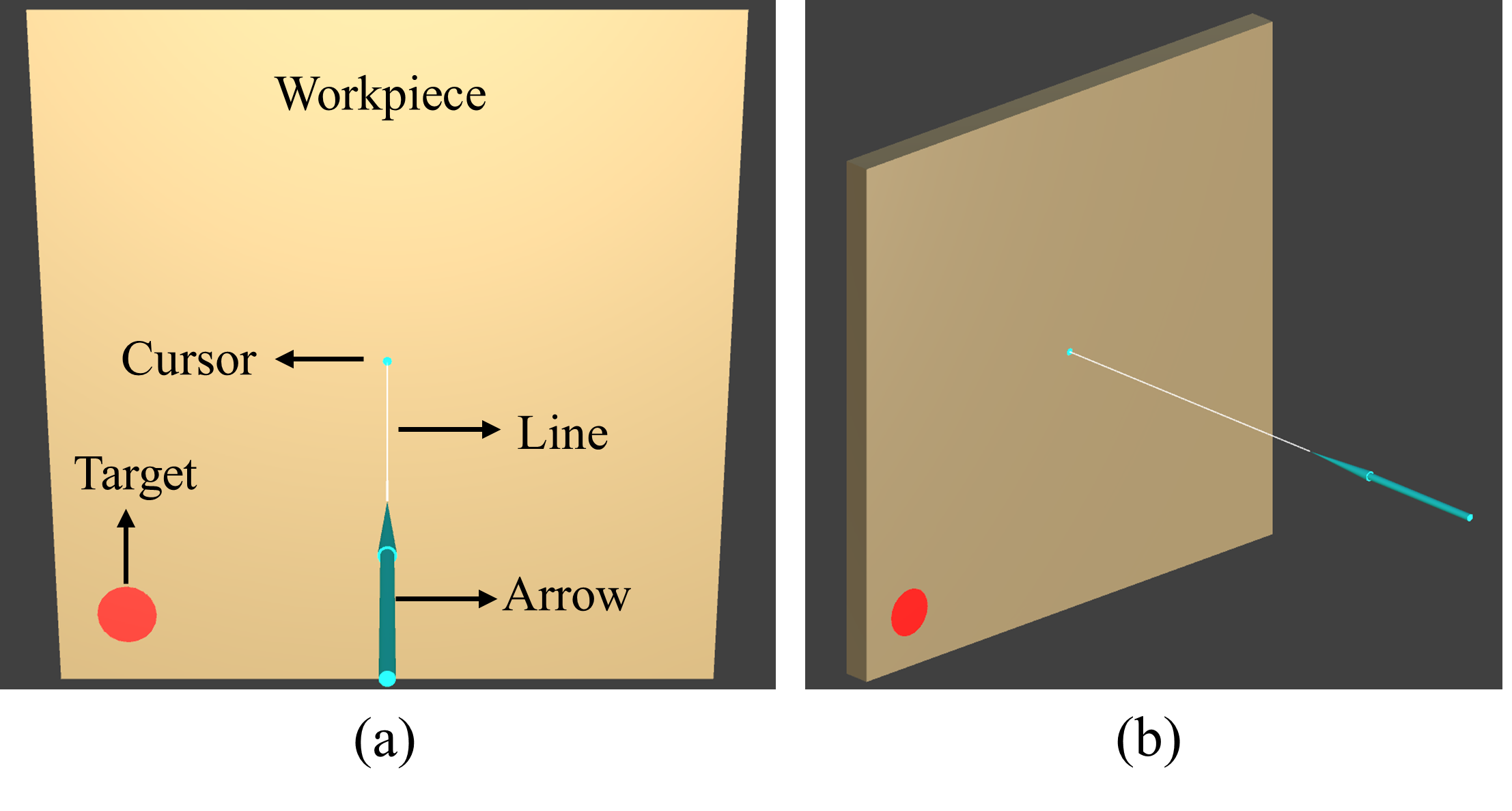}
	\centering
	\caption{Snapshot of the visual feedback displayed to the subjects for the spring compression experiments performed in VEs; (a) What the subjects see on the monitor screen; (b) Another view to aid the imagination.}
    \label{fig:guisnapshot}
\end{figure}

In this experiment, the admittance damping was adapted based on subtask detection but not motion estimation. Since all parameters of the experiment were fully known and under control, an ideal subtask detector was hard-coded into the system, which used predefined and controlled thresholds of human force for detecting \textit{Idle}, \textit{Tool-Attachment}, and \textit{Driving}. The system detected \textit{Contact} as soon as the tip of the virtual drill bit touched the virtual spring, which generated an artificial reaction force.
A snapshot of the visual feedback in these experiments can be seen in Fig.~\ref{fig:guisnapshot}.

\subsection{Protocol}

The following protocol was followed in this experiment:

The subject stood and waited inside the designated area (\textit{Idle}) until the ``GRAB THE HANDLE" command appeared on the screen. They then grabbed the handle of the robot and held it steady (\textit{Tool-Attachment}). After 3 seconds, ``GO" command appeared on the screen, at which point the subject started moving the robot towards the target. As the robot moved, a virtual arrow imitating the drill bit was displayed in real time on the computer screen. The subject was asked to move the arrow in free space (\textit{Driving}) and penetrate the workpiece at the target (\textit{Contact}), imitating the drilling process. The target was represented by a red circle on the workpiece, allowing the user to initiate penetration at any point within the circle. A progress bar appeared when the arrow touched the workpiece anywhere within the circle, showing the current compression depth and target depth (\textit{Contact}). The subject kept penetrating the workpiece until the ``RETRACT" command was shown, officially ending the trial. At this point, the subject retracted the arrow from the workpiece and stood clear for the robot to let it return to its home position and go to the subsequent trial.

\subsection{Conditions and Subjects}

Experimental conditions, summarized in Table~\ref{tab:experiments}, included the following variables:

\begin{itemize}
    \item Perpendicular distance: Normal distance between the home position and the workpiece
    \item Target location: Target was on one of the corners of a 15$\times$15 [cm] square
    \item Target size: Circle diameter; small and large
\end{itemize}

The above conditions were chosen considering that in real pHRI scenarios, target position and size are unknown and can change in every drilling session. To ensure consistency in the durations of \textit{Driving} and human behaviors, and to facilitate meaningful comparisons of target sizes, we drew inspiration from the concept of ``Index of Difficulty" (IoD, expressed in bits) as defined for the Fitts' reaching task~\cite{Soukoreff2004}. As per Shanon's formula~\cite{Soukoreff2004}, for a given IoD, the target diameter can be defined as follows:
\begin{equation}
    W = \frac{L_p}{2^\mathrm{IoD}-1}
    \label{eq:IoD}
\end{equation}
where $W$ [m] is the target diameter, and $L_p$ [m] is the perpendicular distance from the home position to the workpiece. In our experiments, IoD values of 3 and 5 were chosen for large and small targets, respectively. Table~\ref{tab:targetsizes} shows the actual target diameters for each $L_p$ and IoD.

\begin{table}[t]
\myVspace
\centering
\caption{Target diameters $W$ [cm] as a function of Index-of-Difficulty (IoD) and perpendicular distance $L_p$}
\label{tab:targetsizes}
\begin{tabular}{cc|c|c}
\multicolumn{2}{c|}{\textbf{IoD}}                                            & \textbf{3} & \textbf{5} \\ \hline
\multirow{4}{*}{\rotatebox[origin=c]{90}{\textbf{$L_p$ [cm]}}} & \textbf{12} & 1.71       & 0.39       \\
 & \textbf{16} & 2.29 & 0.52 \\
 & \textbf{20} & 2.86 & 0.65 \\
 & \textbf{24} & 3.43 & 0.77
\end{tabular}%
\end{table}

5 subjects (1 female and 4 males; average age: 26.25 ± 4.2 SD) participated in the experiment. There were 32 experimental conditions (4  distances $\times$ 4 corners $\times$ 2 IoD), and all subjects repeated each condition 5 times. Hence, the number of trials for each subject was 160, while the total was 800.

\subsection{Cross Validation}

Structured k-fold cross-validation (CV) was employed during the offline training of the models to evaluate their performance in unforeseen circumstances. In each fold, trials belonging to a single subject, perpendicular distance, target location (corner), or target size (IoD) were kept for testing, while the rest were used for training. This sort of CV showed the expected performance of the models when encountering new subjects or conditions unseen in the training set, which is important in real pHRI scenarios.

\begin{figure}[t]
	\includegraphics[width=\columnwidth]{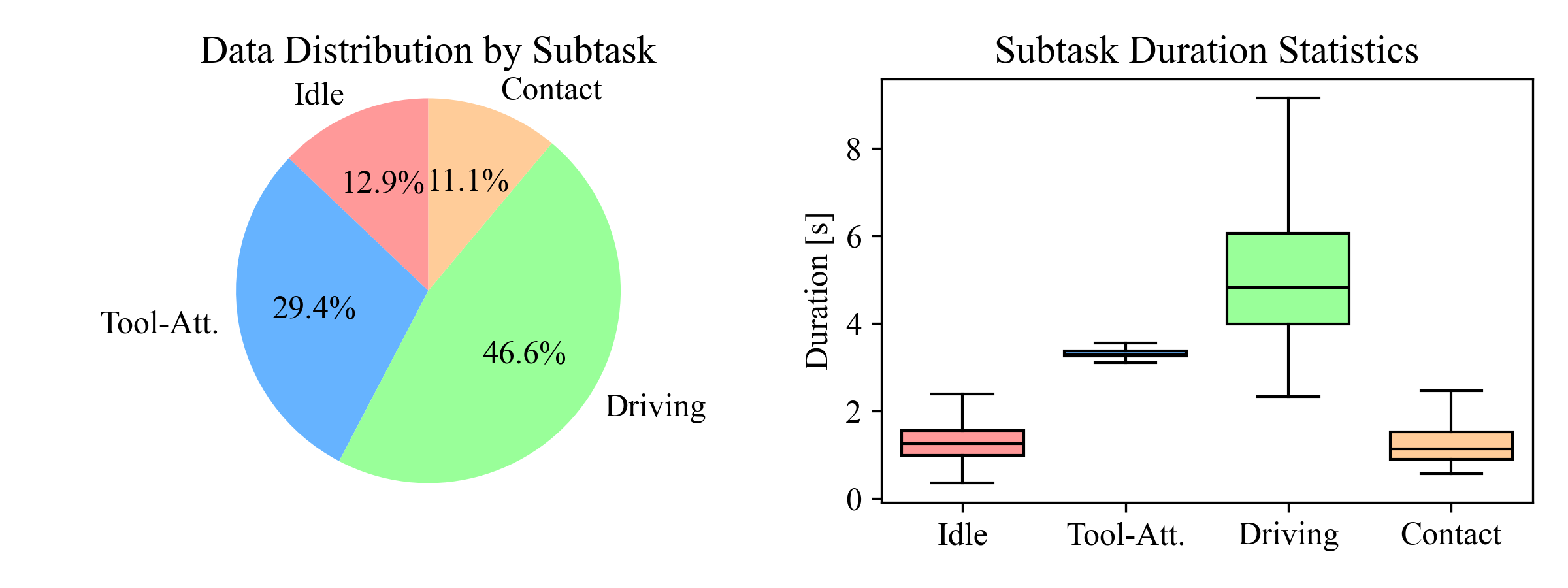}
	\centering
	\caption{Left: Data portion distribution per subtask; Right: Subtask duration statistics, in experiment A}
    \label{fig:subtaskdurations}
\end{figure}

\begin{figure}[t]
	\includegraphics[width=\columnwidth]{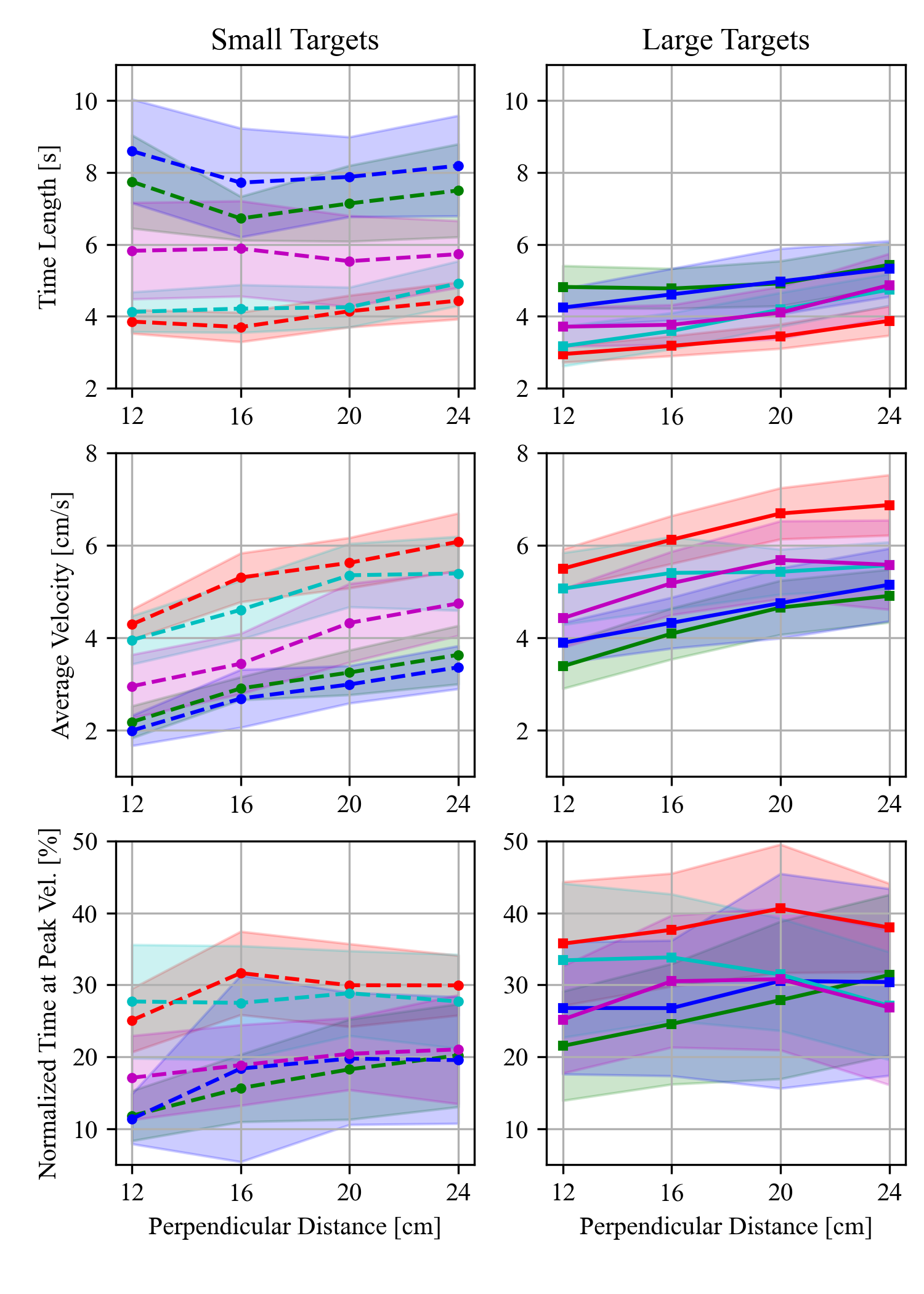}
	\centering
	\caption{Statistics of the Driving phase, including time duration (top row), average velocity (middle row), and time progress at the peak velocity (bottom row). The left column is for the small targets (higher IoD), and the right column is for the large targets (lower IoD). Every color represents a subject. Solid lines are means, while the shaded areas are standard deviations.}
	\label{fig:drivingstats}
\end{figure}

\begin{figure}[t]
	\includegraphics[width=\columnwidth]{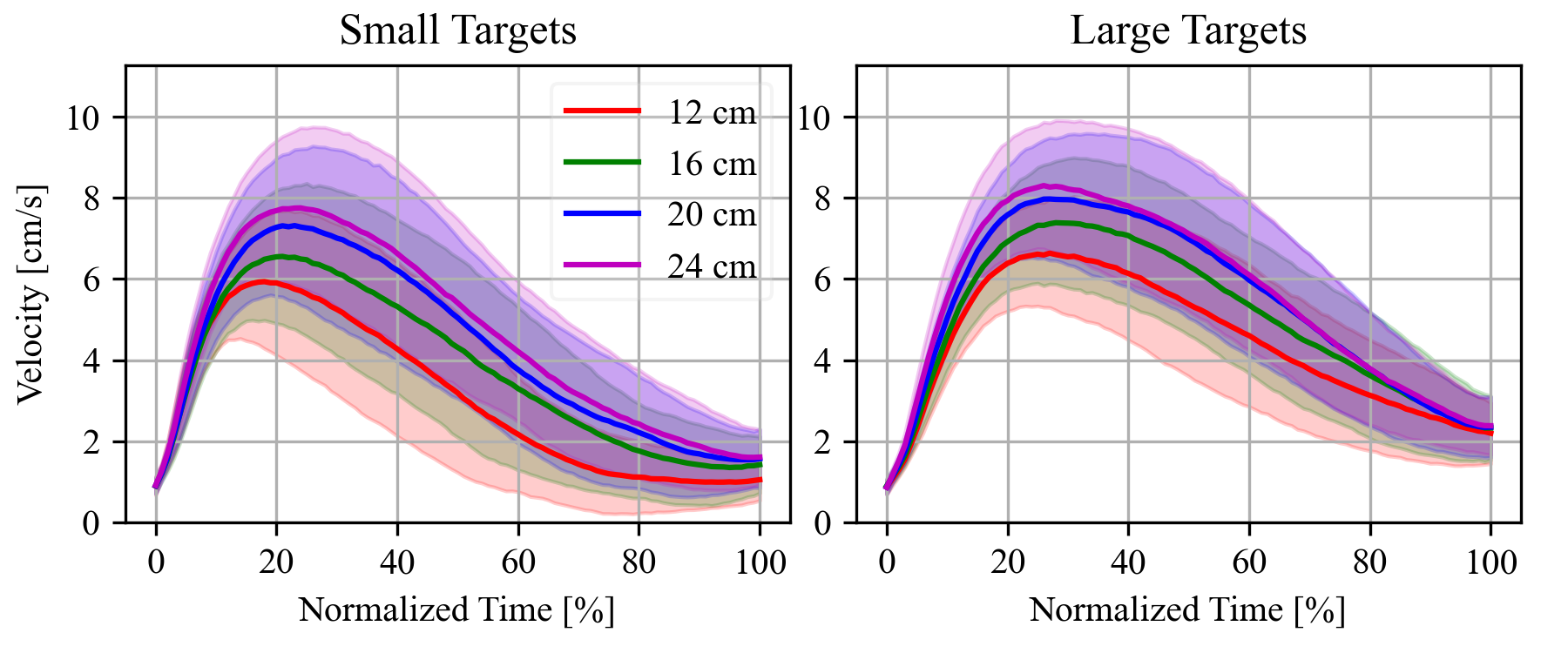}
	\centering
	\caption{Time-normalized profiles of velocity magnitude in experiment A. Every color is a perpendicular distance. Solid curves show means, and shaded areas show the standard deviations.}
    \label{fig:velprofile}
\end{figure}

\begin{figure}[t]
	\includegraphics[width=\columnwidth]{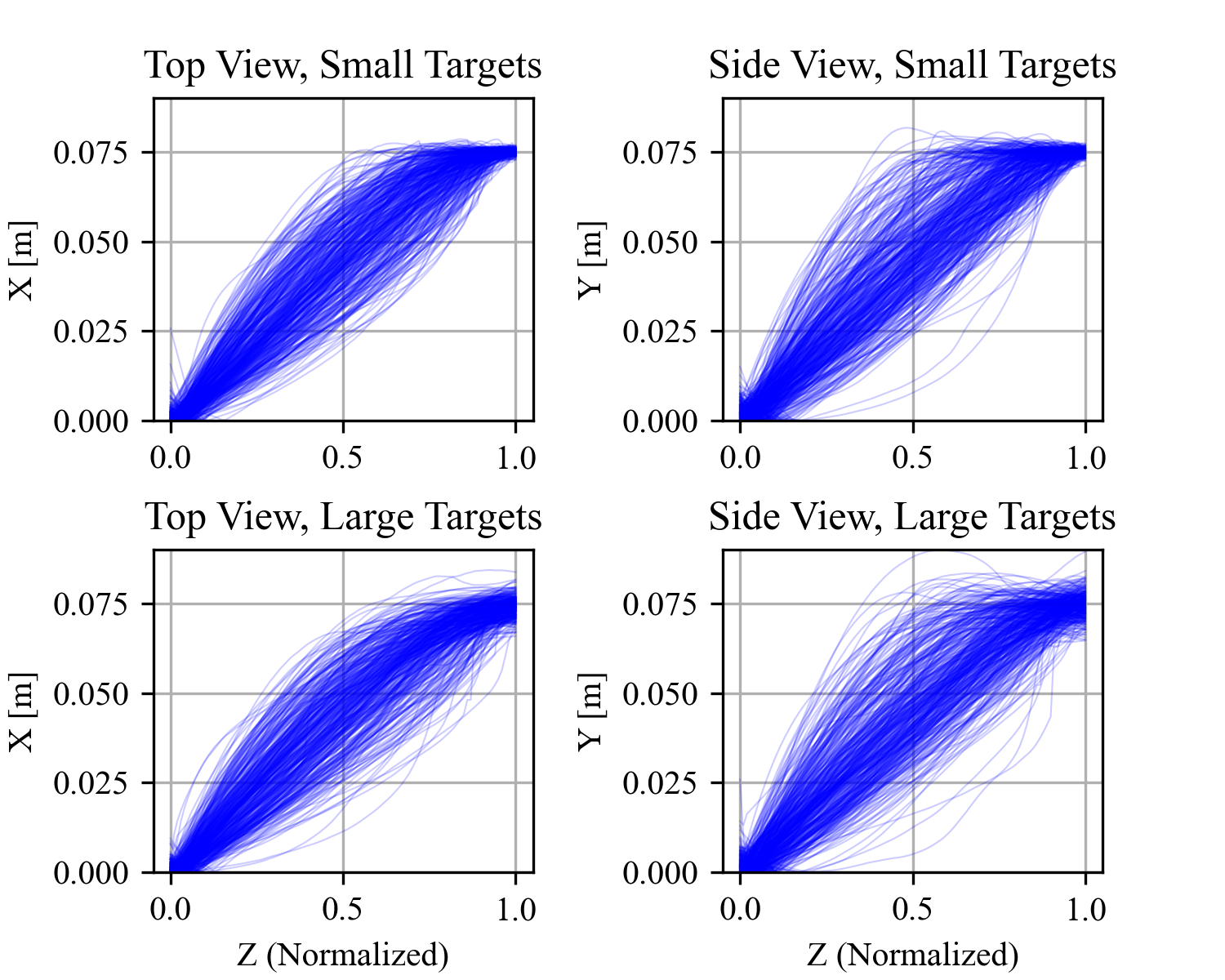}
	\centering
	\caption{Cartesian trajectories of all subjects in the \textit{Driving} phase. The perpendicular distance has been normalized, and negative values have been mirrored to fit all the trials together.}
    \label{fig:cartesiantraj}
\end{figure}

\subsection{Visualizing Training Data}

Fig.~\ref{fig:subtaskdurations} shows the distribution of the four subtasks in the training data together with their durations.
Fig.~\ref{fig:drivingstats} shows the statistics for \textit{Driving} in these experiments. The distribution of time length $T_d$ for \textit{Driving} can be seen on the top row of Fig.\ref{fig:drivingstats}. Accordingly, we observe some differences in the movement behaviors of subjects, even though for all subjects, larger targets (IoD = 3) took a shorter time to drive and align the arrow than smaller targets (IoD = 5). The difference between the two IoDs and the variance of \textit{Driving} time length $T_d$ in every perpendicular distance $L_p$ was different among subjects. The middle row of Fig.~\ref{fig:drivingstats} shows the distribution of average velocities during \textit{Driving}. Accordingly, longer $L_p$ produced higher average velocity, and the relationship between $L_p$ and average velocity was sometimes nonlinear. Average velocity was also always lower with small targets than large targets. The bottom row of Fig.~\ref{fig:drivingstats} shows the distribution of time progress $\tau$ in the \textit{Driving} subtask at peak velocity. As can be seen, there is no visible pattern, meaning data of the initial acceleration phase of \textit{Driving} cannot easily predict when/where the trial will end. Subjects can have sharper acceleration but smoother deceleration than one another, or vice-versa. Fig.~\ref{fig:velprofile} shows the average profile of velocity magnitude in \textit{Driving}. As can be seen, not only ranges of velocities are large in all circumstances throughout \textit{Driving}, but also, there is a variance in the velocity magnitude at \textit{Contact} and just before it, meaning predefined thresholds of variables such as velocity magnitude are not sufficient to estimate when \textit{Contact} is imminent.
Fig.~\ref{fig:cartesiantraj} shows Cartesian trajectories of all subjects. As can be seen, there are seldom any straight-line motions, even though this was a relatively simple pHRI task. The variability in human behaviors under various circumstances is a testament to its inherent complexity and why simple rule-based algorithms often fail when conditions change. 
A conclusion from the figures in this section is that target size (IoD), perpendicular distance $L_p$, and the human subject are the most important features that create the largest variability in the data.

\begin{figure}[t]
	\includegraphics[width=\columnwidth]{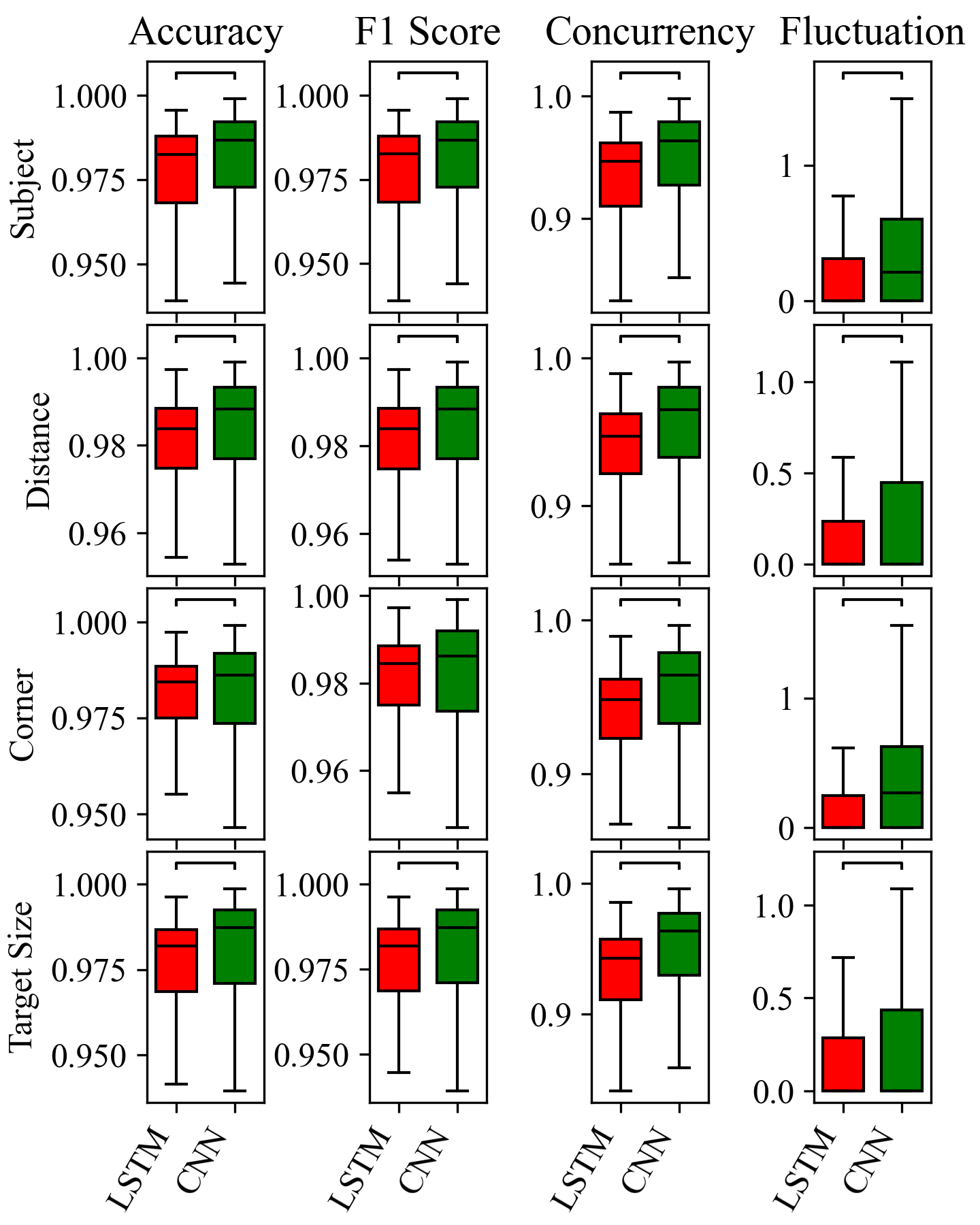}
	\centering
	\caption{Performance of 1D-CNN and LSTM models in subtask classification. Every row is a cross-validation category, and every column is an evaluation metric. In each plot, horizontal brackets denote statistically significant pairwise comparisons with $\alpha=0.001$ following a Bonferroni-corrected Wilcoxon signed rank test.}
    \label{fig:modelstatsstc}
\end{figure}

\subsection{Subtask Detection Results}
\label{sec:expASDResults}

Fig.~\ref{fig:modelstatsstc} compares the performances of 1D-CNN and LSTM models in subtask classification. This figure shows that although both models achieve acceptable performances, the CNN model achieves slightly better accuracy, F\textsubscript{1} score, and concurrency. However, the CNN model also has wider ranges (lower minimums) in some ratings. Furthermore, it has a significantly higher fluctuation rate than LSTM, which is a vital metric of the subtask detector. It is also worth noting that the LSTM model is computationally lighter than the CNN model (Fig.~\ref{fig:dlarch}). Due to these provisions, the LSTM model was chosen in this study for subtask detection.

\begin{figure}[t]
	\includegraphics[width=\columnwidth]{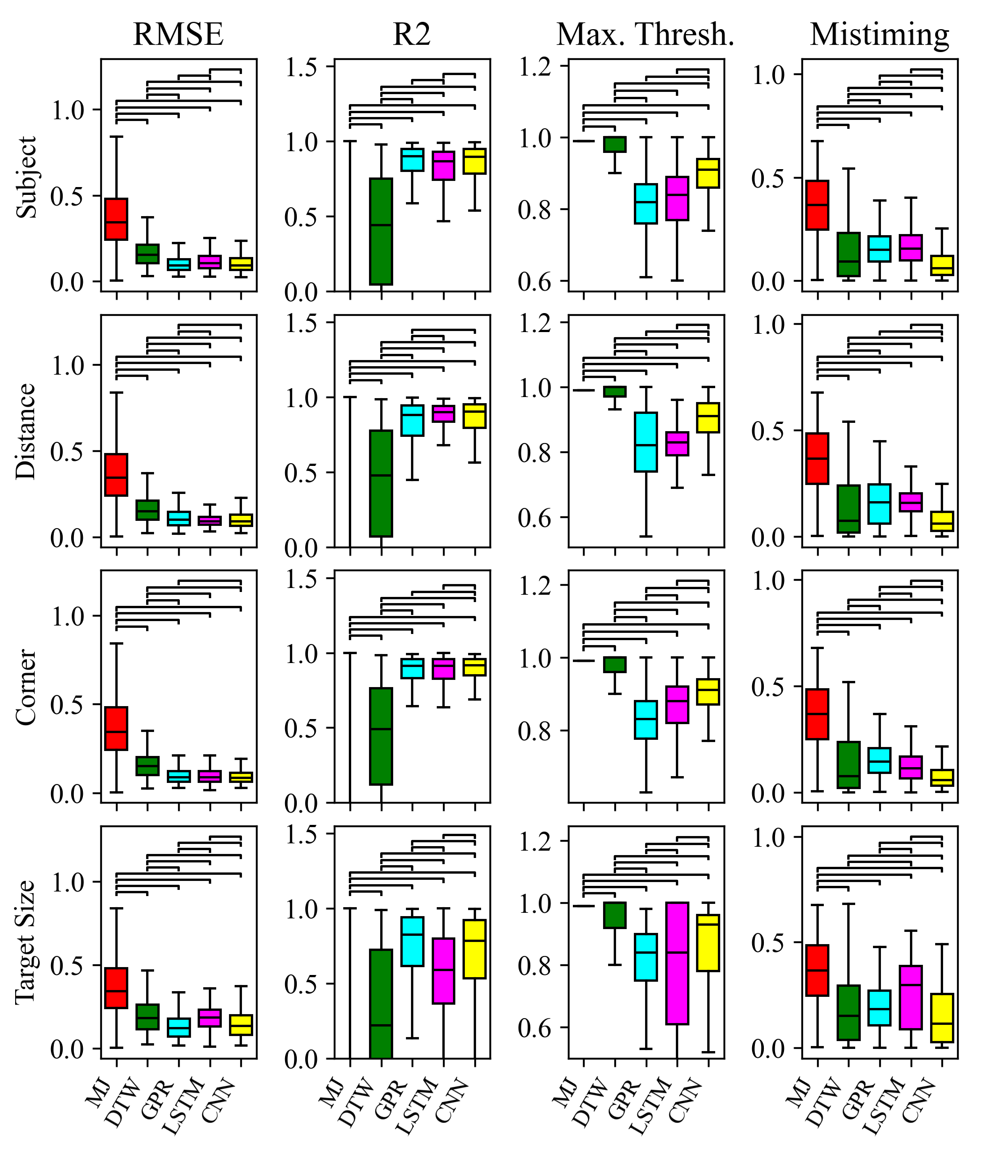}
	\centering
	\caption{Performance of different models in time progress estimation. Every row is a cross-validation category, and every column is an evaluation metric. In each plot, horizontal brackets denote statistically significant pairwise comparisons with $\alpha=0.001$ following a Bonferroni-corrected Wilcoxon signed rank test.}
    \label{fig:modelstatstimeprogest}
\end{figure}

\begin{figure}[t]
	\includegraphics[width=\columnwidth]{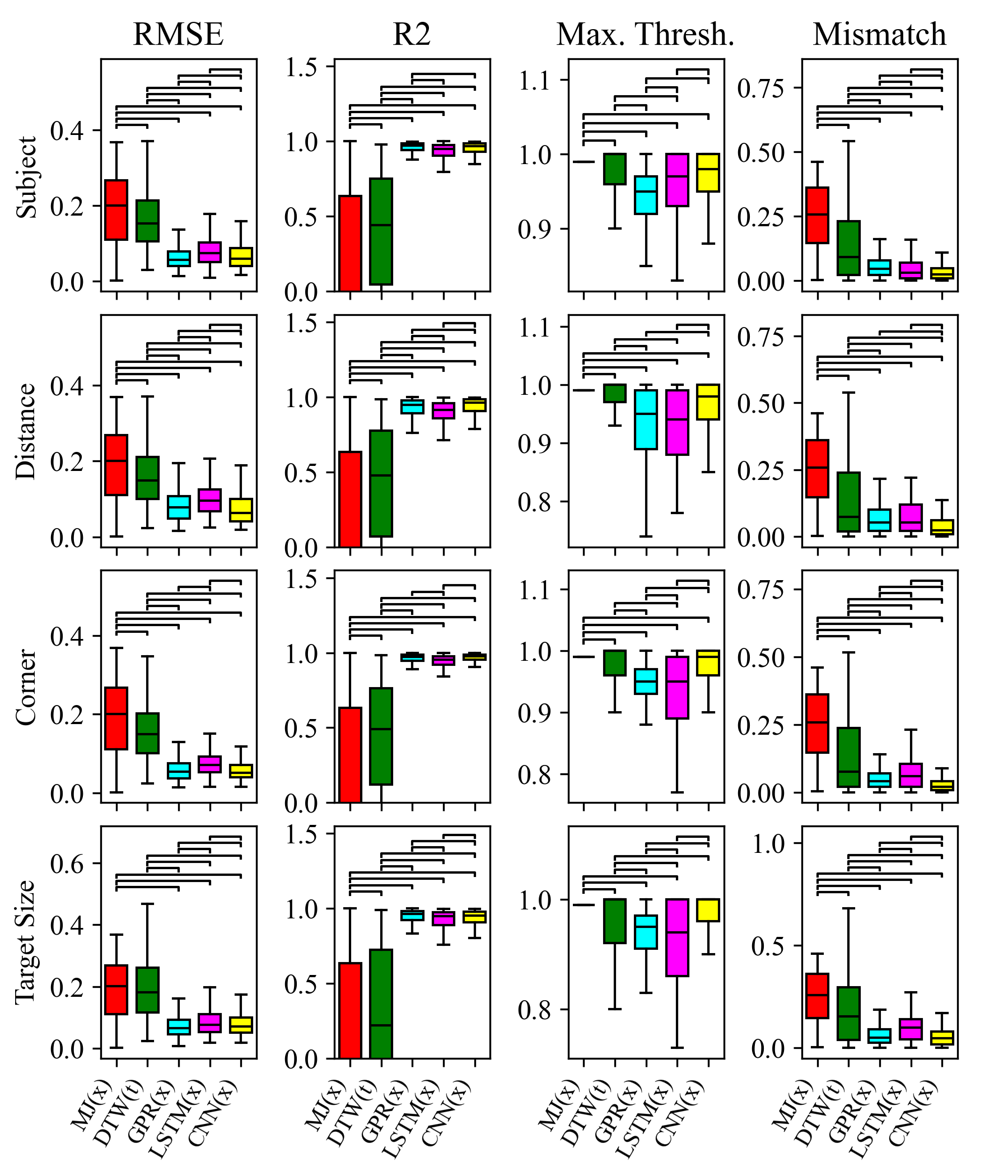}
	\centering
	\caption{Performance of different models in trajectory progress estimation, similar to Fig.~\ref{fig:modelstatstimeprogest}. The suffix `x' denotes trajectory progress estimation, while `t' denotes time progress estimation. Note that, unlike all others, DTW can only be used for time progress estimation.
    }
    \label{fig:modelstatstrajprogest}
\end{figure}

\subsection{Motion Estimation Results}
\label{sec:expAMEResults}

Fig.~\ref{fig:modelstatstimeprogest} and Fig.~\ref{fig:modelstatstrajprogest} show the performance of all models in all CV categories for time and trajectory progress estimation, respectively. The Minimum-Jerk model is the only one with no training/testing set as it is only used at inference time. It should be noted that the metrics \textit{Maximum Adaptation Threshold} and \textit{Mistiming/Mismatch} should be considered equally important, as a model which reaches a progress value of 100\% very early would have a Maximum Threshold of 100\%, but very high Mistiming/Mismatch values, and a model which only reaches a low progress value at \textit{Contact}, would have nearly zero Mistiming/Mismatch, but a very low Maximum Threshold. It can be understood from these figures that overall, GPR, LSTM, and CNN have good performances, with CNN having the best performance, both in time and trajectory progress estimation. It can also be seen that the trajectory progress estimator almost always shows better performance than the time progress estimator. The metrics observed in these figures are the performances for the test set, i.e., performance over testing trials in all folds of the k-fold CV. Since Fig.~\ref{fig:velprofile} and Fig.~\ref{fig:cartesiantraj} show that the trajectories in our experiments are far from ideal symmetric minimum-jerk profiles, the subpar performance of the minimum-jerk model can be observed in Fig.~\ref{fig:modelstatstimeprogest} and Fig.~\ref{fig:modelstatstrajprogest}. As for the DTW method, since large variances exist in the data according to Fig.~\ref{fig:drivingstats} and Fig.~\ref{fig:velprofile}, the templates were not sufficiently representative of the motions, and failed to capture their stochasticity across subjects and conditions. Paired with the inherently subpar performance of DTW in partial (incomplete) matching, these shortcomings resulted in weak performances for the DTW method, as observed in Fig.~\ref{fig:modelstatstimeprogest} and Fig.~\ref{fig:modelstatstrajprogest}. The best overall performance appears to belong to 1D CNN architecture, with trajectory progress estimation.




\section{Experiment B: Online Testing in VEs}
    \label{sec:expB}

\subsection{Protocol, Conditions and Subjects}

These experiments were conducted within an identical virtual environment (VE), utilizing the same scenario and protocol as Experiment A. However, adaptation was based on the trained subtask detector and motion estimator deployed in the system. Also, numerical values for some conditions differed from the training experiments (Experiment A) to introduce new conditions unseen during training (Table~\ref{tab:experiments}). Accordingly, while keeping the four corners of the workpiece, only one perpendicular distance (18 cm) and one target size with IoD = 4 were chosen, both unseen by the trained model.
In these experiments, three admittance controllers were used: (\textbf{C1}) A fixed controller with a constant damping $b=500$~[Ns/m], (\textbf{C2}) An adaptive controller with the same parameters used in Experiment A, which used learning-based subtask detection alone for damping adaptation, without any motion estimation, and (\textbf{C3}) An adaptive controller similar to the previous one, but also incorporating the motion estimator during  \textit{Driving}, to adapt the damping before \textit{Contact} occurs.

3 subjects (1 female and 2 males; average age: 28.67 ± 4.92 SD) participated in the experiment. There were 12 experimental conditions (3 controllers $\times$ 1 perpendicular distance $\times$ 4 target locations $\times$ 1 IoD), and all subjects repeated each experimental condition twice. Hence, the number of trials for each subject was 24, while the total was 72.

\subsection{Results}

\begin{figure}[t]
    \myVspace
	\includegraphics[width=\columnwidth]{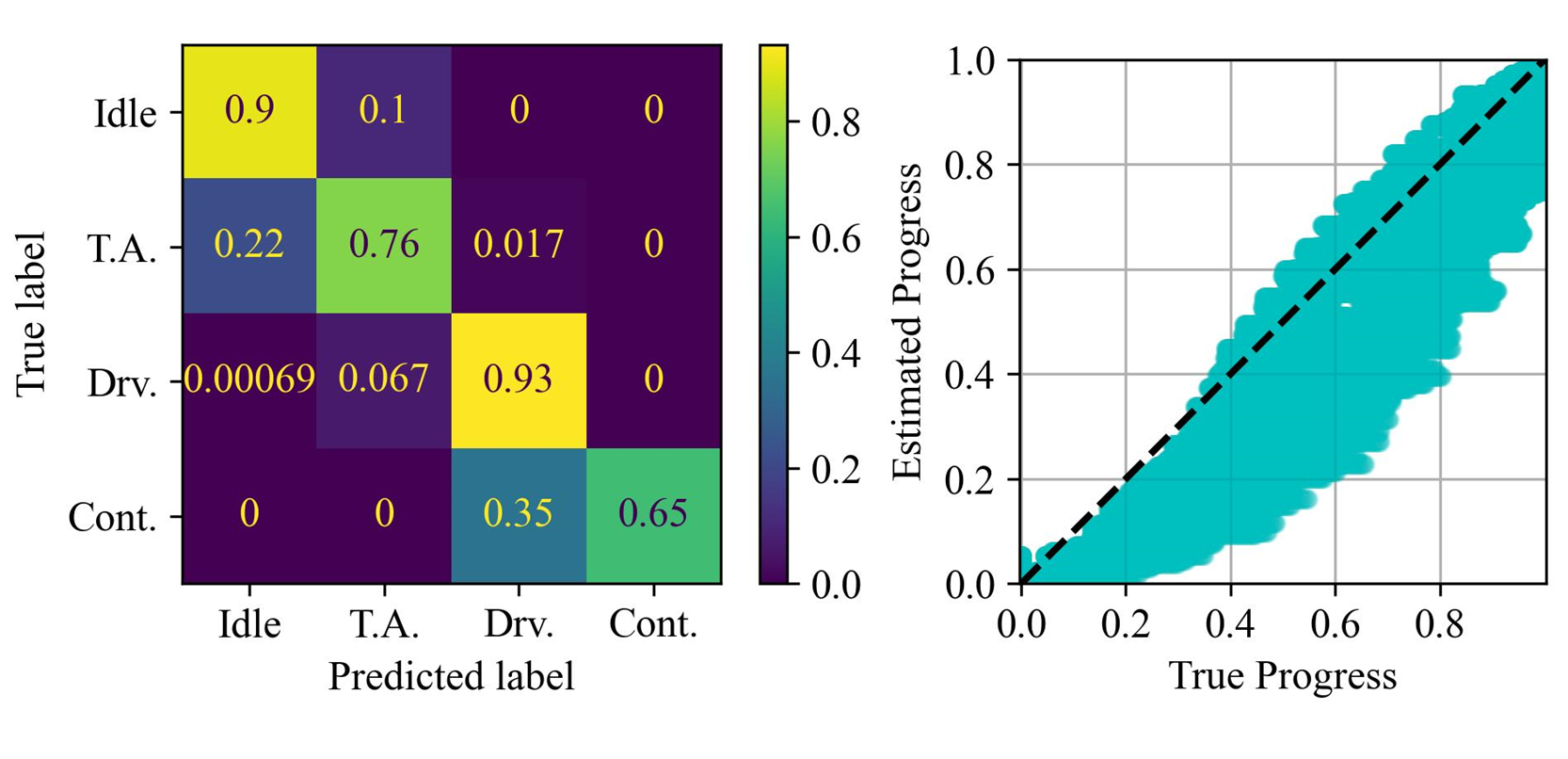}
	\centering
	\caption{The confusion matrix for subtask classification (left) and motion estimation performance (right) in Experiment B (online testing).
    }
    \label{fig:cmRegExpB}
\end{figure}

The performance metrics reported here are extracted from the system's real-time data, collected at 500 Hz. However, the DL models deployed in the system had much lower update rates, inevitably decreasing the real-time subtask detection and motion estimation performances. \pn{This can be alleviated by utilizing GPU or model parallelization.} The subtask detector achieved an accuracy of 84.46\% and a weighted F\textsubscript{1} score of 0.8423, \pn{which is considered satisfactory in most classification tasks.} Most misclassifications by the subtask detector stemmed from delays in detecting subtasks during subtask transitions. This delay is attributed partly to the time needed to accumulate sufficient data from a new subtask and partly to the model's low update rate in real time.
The motion estimator was deployed on the system and activated promptly upon detecting the \textit{Driving} subtask by the subtask detector. A conservative adaptation threshold of 75\% was chosen to adapt the damping before \textit{Contact} occurred. Early adaptation of damping was successfully achieved in all trials under controller C3. The motion estimator achieved an overall RMSE of 0.0890 and R\textsuperscript{2} score of 0.9562, \pn{which are acceptable for our pHRI application}. Despite encountering limitations during their online deployment in the real-time system, the subtask detector and motion estimator demonstrated satisfactory performance, allowing for the appropriate adaptation of the admittance controller. Fig.~\ref{fig:cmRegExpB} shows the confusion matrix for the subtask detector, as well as the overall regression performance of the motion estimator. \pn{Comparable model performances between Exp. B and Exp. A (Sections~\ref{sec:expASDResults},~\ref{sec:expAMEResults}) despite unforeseen conditions showcase the generalization capability of the models.}

\begin{figure}[t]
	\includegraphics[width=\columnwidth]{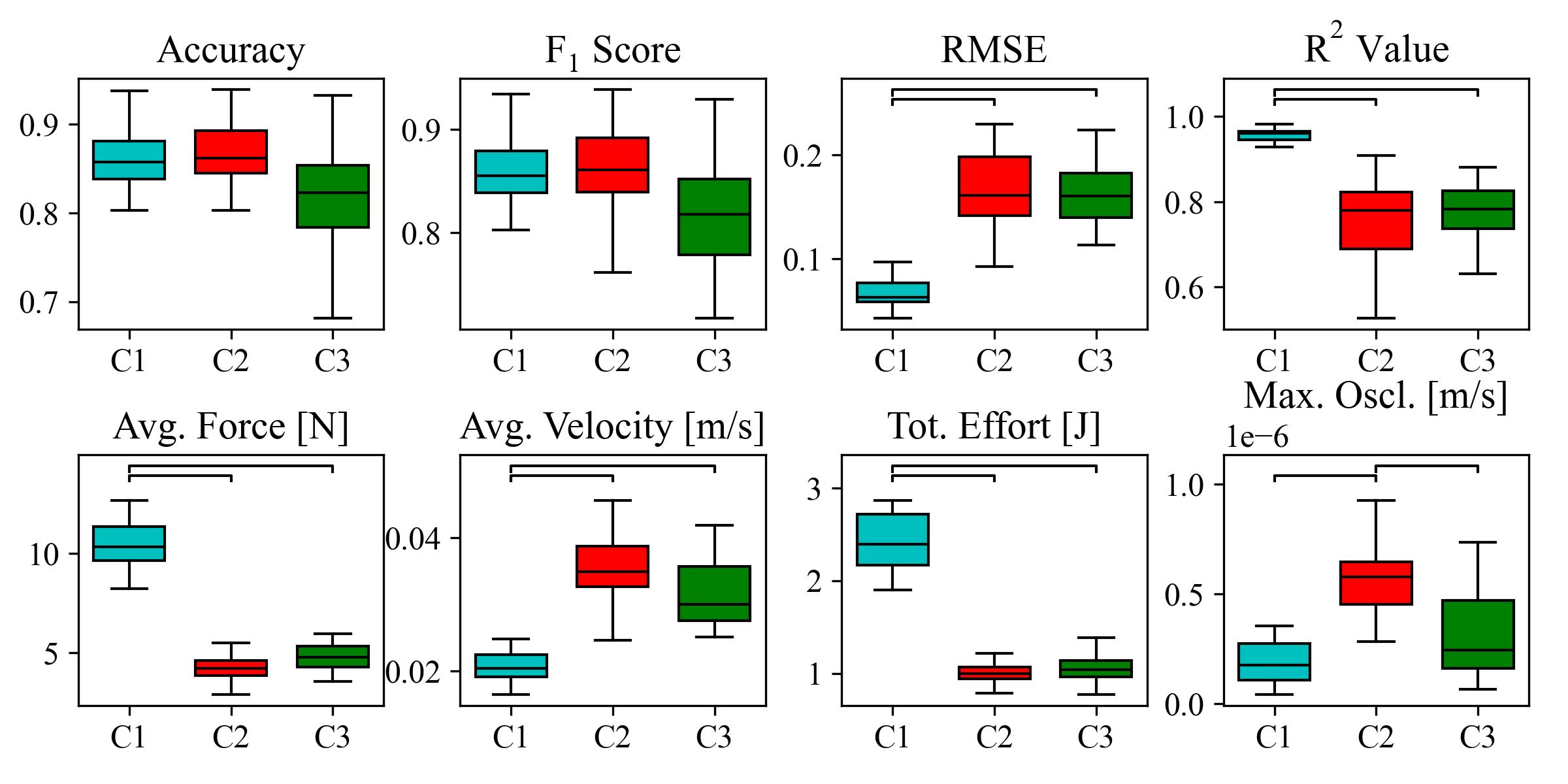}
	\centering
	\caption{Performance metrics observed in Experiment B (online testing). On the top row, the left pair of plots denotes subtask classification metrics, while the right pair denotes motion estimation metrics. The bottom row shows task performance metrics. Box plots show medians and quartiles. Horizontal brackets show statistically significant pairwise comparisons with $\alpha=0.001$ after performing a Bonferoni-corrected Wilcoxon signed rank test.
    }
    \vspace{-1.5em}
    \label{fig:resultsExpB}
\end{figure}

The comparative statistics of all performance metrics introduced in Section~\ref{sec:evaluationmetrics} are reported in Fig.~\ref{fig:resultsExpB}. 
Accordingly, even though the damping profiles and interaction dynamics differed among the three controllers, there were no statistically significant differences between subtask detection performances. This proves that a well-trained learning-based subtask detector will be sufficiently robust to changes in damping, adaptation mechanism, and kinematic/kinetic variables.

The right pair of plots on the top row of Fig.~\ref{fig:resultsExpB} reports motion estimation metrics. As can be seen, some differences exist between C1 (fixed controller) and other adaptive controllers C2 and C3, mainly because C1 incorporates high damping values during the \textit{Driving}, resulting in slower movement, giving more time for the DL model to predict the progress.

The bottom row in Fig.~\ref{fig:resultsExpB} shows adaptive control performance in Experiment B. Accordingly, as was expected, average human force and total human effort were significantly higher in C1 with fixed damping values, while average velocity was significantly lower in C1. This is because of the low damping value during \textit{Driving} in C2 and C3. The lower right plot in this figure shows the peak oscillation amplitude of the end-effector velocity. Accordingly, in C1 and C3, because the \textit{Contact} (when the robot touches the workpiece) began with high damping, there were significantly smaller oscillations than those observed in C2, in which the inevitable delay in subtask detection resulted in low damping at \textit{Contact}. As can be seen from Fig.~\ref{fig:resultsExpB}, when using C3, not only was the human effort and force minimized, but also, instabilities were avoided at the beginning of \textit{Contact}, thereby maximizing stability and safety.
\pn{Small differences in some metrics observed in Fig.~\ref{fig:resultsExpB} between C2 and C3 occurred because in C3 the damping increased to a high value during \textit{Driving} and some time before \textit{Contact}. This short period of time spent at the end of \textit{Driving} with high damping lead to a slight increase in force and effort in \textit{Driving}, though statistically insignificant.}

\begin{figure}[t]
    \myVspace
	\includegraphics[width=\columnwidth]{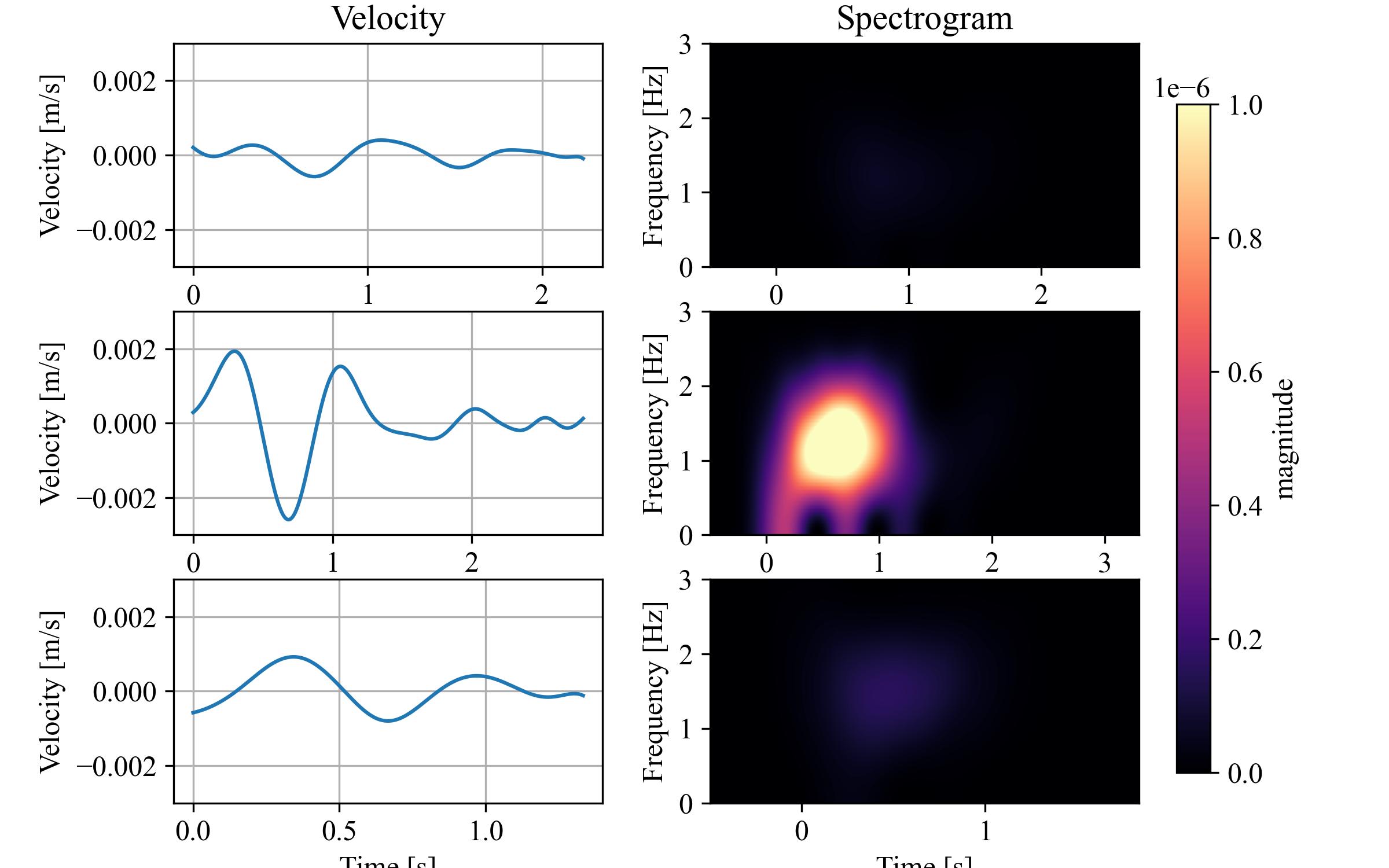}
	\centering
	\caption{Sample trial from Experiment B (online testing) showing oscillations of the end-effector velocity during the \textit{Contact}. The left column shows the end-effector velocity as a function of time, and the right column shows its spectrogram with $\Delta f = 0.05 \text{[Hz]}$. Rows 1 to 3 correspond with C1 to C3, respectively. 
    }
    \vspace{-1.5em}
    \label{fig:spectrogramExpB}
\end{figure}

Fig.~\ref{fig:spectrogramExpB} shows the oscillations of the end-effector velocity for a sample trial of Experiment B. As can be seen, in C1 and C3 where \textit{Contact} began with high damping, oscillations are minimal. In contrast, in C2 (middle row), they are distinctly larger because damping is not adapted appropriately during the initial moments of \textit{Contact}.
In real-life scenarios with rigid materials, these instabilities occurring in C2 can easily damage the sensors in the pHRI system and the workpiece. This demonstrates the benefit of the motion estimator in contact-rich pHRI tasks.

\section{Experiment C: Verification in Physical World}
    \label{sec:expC}

These experiments evaluated whether the proposed models, originally trained in VEs, could be seamlessly transferred to real-world settings without compromising performance. \pn{To that end, model performances should not drop dramatically in these experiments compared to Experiment B.}
These experiments involved collaborative drilling of four corners of a 7$\times$7 [cm] square on a flat 1-cm-thick plywood plate, positioned vertically, using a powered drill attached to the robot's end-effector.

\subsection{Protocol, Conditions, and Subjects}

The drilling experiments in the physical world followed a protocol and conditions similar to Experiment B, except that the screen was solely utilized for providing instructions and visual feedback. Furthermore, only the C3 configuration (adaptive control with subtask detection and motion estimation) was employed (see Table~\ref{tab:experiments}). The same subjects as Experiment B were included. Model performances were evaluated after the experiments.

\subsection{Results}

\begin{figure}[t]
    \myVspace
	\includegraphics[width=\columnwidth]{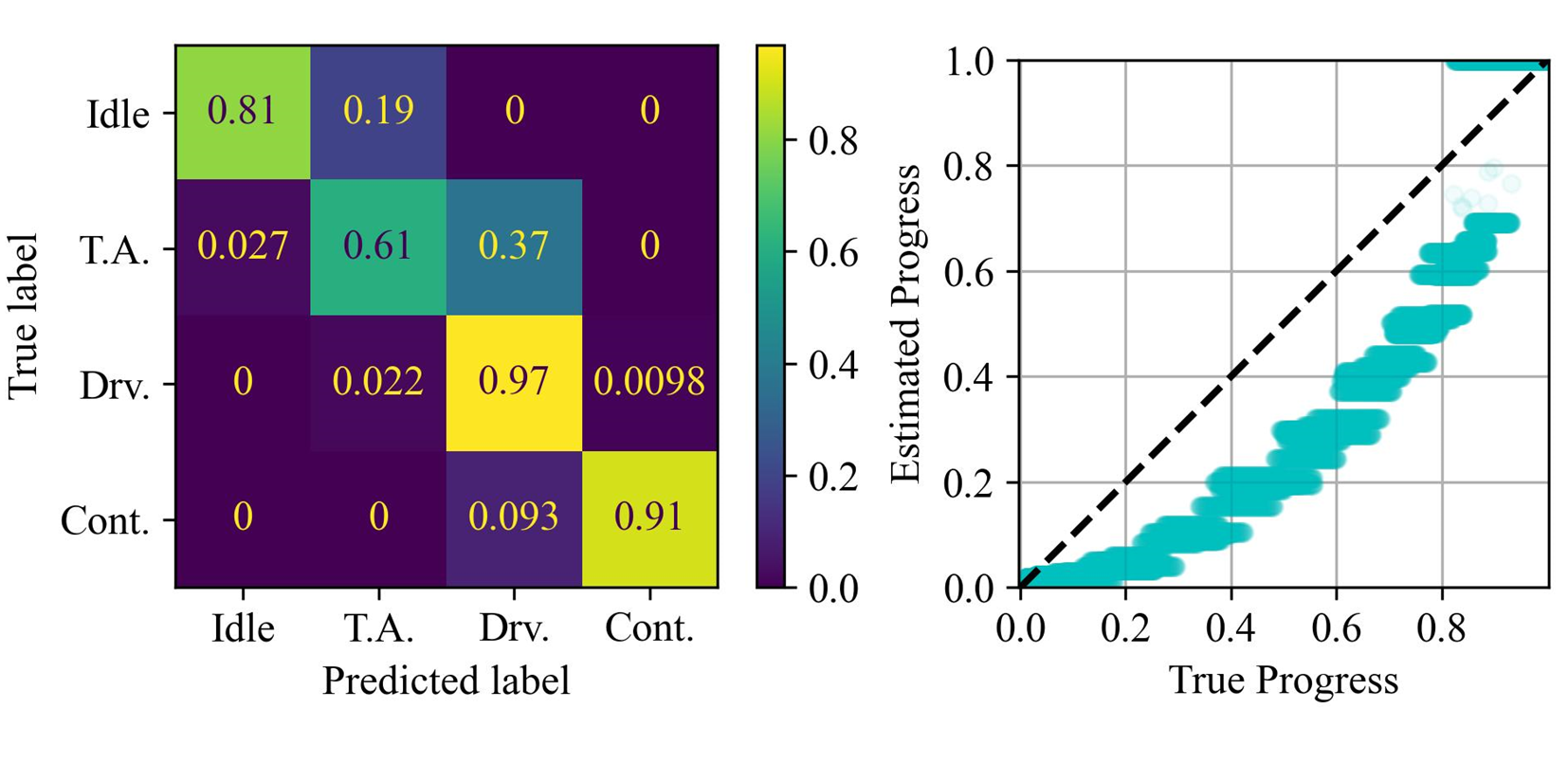}
	\centering
	\caption{The confusion matrix for subtask classification (left) and motion estimation performance (right) in Experiment C (real drilling in the physical world).
    }
    \vspace{-1.5em}
    \label{fig:cmRegExpC}
\end{figure}

In the drilling experiments performed in the physical world, the system achieved a subtask classification accuracy of 83.61\%, and a weighted F\textsubscript{1} Score of 0.8361, along with a motion estimation RMSE of 0.0948 and an R\textsuperscript{2} Score of 0.96. \pn{These values, similar to those reported in Experiment B, are often considered adequate in most classification and regression tasks.} Despite the possible differences in human behavior and/or trajectories between the virtual world and the real world, the performances of the models are sufficiently similar to those observed in Experiment B and adequate for successful adaptation in \textit{Driving} and just before \textit{Contact}. The performances achieved by the subtask detector and motion estimator are depicted in Fig.~\ref{fig:cmRegExpC}.

\section{Discussion and Conclusion}
    \label{sec:conclusion}
    
This study emphasizes that many contact-rich pHRI tasks in small-batch manufacturing, such as drilling, grinding, polishing, cutting, etc., can be deconstructed into subtasks, e.g., \emph{Idle}, \emph{Tool-Attachment}, \emph{Driving}, and \emph{Contact}. We define human intention as the specific subtask currently in progress and how much of the \textit{Driving} subtask is progressed, which defines human motion intention more precisely and helps with handling possible \textit{Contact} instabilities. 
We argue that if a robot can recognize the subtask and estimate motion progress effectively, it can adapt its controller accordingly to improve both task efficiency and contact safety.
To this end, we implemented subtask classification utilizing kinematic and kinetic signals. Following subtask detection, the admittance damping was adjusted to comply with human intention.
Starting from an intermediate value to eliminate any jerky movements at the \emph{Tool-Attachment}, a low admittance damping was used during the \emph{Driving} to improve transparency (minimal resistance to human movement), while a high value was used during the \emph{Contact} for stable interactions.
One of the primary challenges we faced was the timely detection of \textit{Contact} before it arose to mitigate potential stability issues, including oscillation or sudden bounce-back. To address this challenge, we introduced a motion estimator that allowed us to anticipate and proactively adapt the controller before the \emph{Contact} subtask was initiated.

We conducted 3D experiments in virtual and physical worlds to validate our approach, focusing on collaborative drilling.
Here, it is important to emphasize the benefits of virtual worlds in training our ML model, which is in line with the concept of simulation to reality, \emph{Sim2Real}.
The virtual environment enabled us to conduct simulated drilling experiments under different experimental conditions, reducing the number of trials for training the ML model in the physical world.
Moreover, since the control loop shown in Fig.~\ref{fig:controlsystem} runs at a high update rate of 500 Hz, real-time haptic rendering of contact forces was possible in our pHRI system. 
Our experimental results showed that while subtask detection leads to significantly lower human effort and faster task execution, motion estimation allows the anticipation of \textit{Contact} and early adaptation of the controller, leading to significantly smaller oscillations and better contact stability.

An intriguing avenue for future research involves the application of unsupervised time series segmentation to partition the contact-rich pHRI tasks into subtasks.
Conversely, in our current study, we operated under the assumption that the ideal admittance damping for each subtask were pre-defined. However, the challenge of selecting suitable damping parameters for individual subtasks remains an open research question that we aspire to address in the future. 
Another compelling research direction could entail dynamic adjustments to damping parameters tailored to the specific performance demands of each subtask. Reinforcement learning (RL) techniques can be leveraged to devise optimal policies that address these specific requirements more effectively.


%



\section*{Acknowledgment}
P.P.N. acknowledges the research fellowship provided by the KUIS AI Center.

\ifCLASSOPTIONcaptionsoff
  \newpage
\fi

\bibliography{references}

\end{document}